\documentclass{article}

\usepackage[final]{neurips_2019}

\usepackage[utf8]{inputenc} % allow utf-8 input
\usepackage[T1]{fontenc}    % use 8-bit T1 fonts
\usepackage{hyperref}       % hyperlinks
\usepackage{url}            % simple URL typesetting
\usepackage{booktabs}       % professional-quality tables
\usepackage{amsfonts}       % blackboard math symbols
\usepackage{nicefrac}       % compact symbols for 1/2, etc.
\usepackage{microtype}      % microtypography
\usepackage{graphicx}
\usepackage{wrapfig}
\usepackage{lipsum}
\usepackage{amsmath}

\newtheorem{theorem}{Theorem}

\newcommand{\tQ}{\widetilde{Q}}

\newcommand\spaceSec{0.0} 
\newcommand\spaceSubSec{0.0}

\DeclareMathOperator*{\argmax}{arg\,max}

\title{Using a Logarithmic Mapping to Enable Lower Discount Factors in Reinforcement Learning}

\author{%
  Harm van Seijen \\
  Microsoft Research Montr\'eal \\
  \texttt{harm.vanseijen@microsoft.com} \\
  \And
   Mehdi Fatemi \\
   Microsoft Research Montr\'eal \\
   \texttt{mehdi.fatemi@microsoft.com} \\
   \AND
   Arash Tavakoli \\
   Imperial College London \\
   \texttt{a.tavakoli@imperial.ac.uk} \\
}

\begin{document}

\maketitle

\begin{abstract}
In an effort to better understand the different ways in which the discount factor affects the optimization process in reinforcement learning, we designed a set of experiments to study each effect in isolation. Our analysis reveals that the common perception that poor performance of low discount factors is caused by (too) small action-gaps requires revision. We propose an alternative hypothesis that identifies the \emph{size-difference} of the action-gap across the state-space as the primary cause. We then introduce a new method that enables more homogeneous action-gaps by mapping value estimates to a logarithmic space. We prove convergence for this method under standard assumptions and demonstrate empirically that it indeed enables lower discount factors for approximate reinforcement-learning methods. This in turn allows tackling a class of reinforcement-learning problems that are challenging to solve with traditional methods.
\end{abstract}

\vspace{-\spaceSec em}
\section{Introduction}
\vspace{-\spaceSec em}

In reinforcement learning (RL), the objective that one wants to optimize for is often best described as an undiscounted sum of rewards (e.g., maximizing the total score in a game) and a discount factor is merely introduced so as to avoid some of the optimization challenges that can occur when directly optimizing on an undiscounted objective \citep{Bertsekas:1996:NP:560669}. In this scenario, the discount factor plays the role of a hyper-parameter that can be tuned to obtain a better performance on the true objective. Furthermore, for practical reasons, a policy can only be evaluated for a finite amount of time, making the effective performance metric a \emph{finite-horizon, undiscounted objective}.\footnote{As an example, in the seminal work of \cite{mnih:nature15}, the (undiscounted) score of Atari games is reported with a time-limit of 5 minutes per game.}

To gain a better understanding of the interaction between the discount factor and a finite-horizon, undiscounted objective, we designed a number of experiments to study this relation. One surprising finding is that for some problems a low discount factor can result in better asymptotic performance, when a finite-horizon, undiscounted objective is indirectly optimized through the proxy of an \emph{infinite-horizon, discounted sum}. This motivates us to look deeper into the effect of the discount factor on the optimization process. 

We analyze why in practice the performance of low discount factors tends to fall flat when combined with function approximation, especially in tasks with long horizons. Specifically, we refute a number of common hypotheses and present a new one instead, identifying the primary culprit to be the size-difference of the action gap (i.e., the difference between the values of the best and the second-best actions of a state) across the state-space.

Our main contribution is a new method that yields more homogeneous action-gap sizes for sparse-reward problems. This is achieved by mapping the update target to a logarithmic space and performing updates in that space instead. We prove convergence of this method under standard conditions. 

Finally, we demonstrate empirically that our method achieves much better performance for low discount factors than previously possible, providing supporting evidence for our new hypothesis. Combining this with our analytical result that there exist tasks where low discount factors outperform higher ones asymptotically suggests that our method can unlock a performance on certain problems that is not achievable by contemporary RL methods.

\vspace{-\spaceSec em}
\section{Problem Setting}
\vspace{-\spaceSec em}

%In the section, we introduce the basic problem setting. We also introduce a few definitions that allow us to make more precise statements about the discount factor.

Consider a Markov decision process (MDP, \citep{puterman:markov94}) $M = \langle \mathcal{S}, \mathcal{A}, P, R, S_0 \rangle$, where $\mathcal{S}$ denotes the set of states, $\mathcal{A}$ the set of actions, $R$ the reward function $R:  \mathcal{S} \times \mathcal{A} \times\mathcal{S}  \rightarrow \mathbb{R}$, $P$ the transition probability function $P : \mathcal{S} \times \mathcal{A} \times \mathcal{S} \rightarrow [0, 1]$, and $S_0$ the starting state distribution. At each time step $t$, the agent observes state $s_t \in \mathcal{S}$ and takes action $a_t \in \mathcal{A}$. The agent observes the next state $s_{t+1}$, drawn from the transition probability distribution $P(s_t, a_t, \cdot)$, and a reward $r_{t} = R(s_t, a_t, s_{t+1})$. A $\emph{terminal state}$ is one that, once entered, terminates the interaction with the environment; mathematically, it can be interpreted %as a state with a single action pointing to itself  
as an absorbing state that transitions only to itself
with a corresponding reward of 0.
The behavior of an agent is defined by a policy $\pi$, which, at time step $t$, takes as input the history of states, actions, and rewards, $s_0, a_0, r_0, s_1, a_1, .... r_{t-1}, s_t$, and outputs a distribution over actions, in accordance to which action $a_t$ is selected. If action $a_t$ only depends on the current state $s_t$, we will call the policy a \emph{stationary} one; if the policy depends on more than the current state $s_t$, we will call the policy \emph{non-stationary}.

We define a \emph{task} to be the combination of an MDP $M$ and a \emph{performance metric} $F$. The metric $F$ is a function that takes as input a policy $\pi$ and outputs a score that represents the performance of $\pi$ on $M$. By contrast, we define the \emph{learning metric} $F_l$ to be the metric that the agent optimizes. Within the context of this paper, unless otherwise stated, the performance metric $F$ considers the expected, finite-horizon, undiscounted sum of rewards
over the start-state distribution; the learning metric $F_l$ considers the expected, infinite-horizon, discounted sum of rewards:
\begin{equation}
F(\pi, M)= \mathbb{E}\left[ \sum_{i=0}^{h-1} r_i \Big| \pi, M \right] \qquad ; \qquad
F_l(\pi, M)= \mathbb{E}\left[ \sum_{i=0}^\infty \gamma^{i} r_i \Big| \pi, M \right]\,,
\label{eq:metrics}
\end{equation}
where the horizon $h$ and the discount factor $\gamma$ are hyper-parameters of $F$ and $F_l$, respectively.

The optimal policy of a task, $\pi^*$, is the policy that maximizes the metric $F$ on the MDP $M$. Note that in general $\pi^*$ will be a non-stationary policy. In particular, the optimal policy depends besides the current state on the time step. We denote the policy that is optimal w.r.t. the learning metric $F_l$ by $\pi^*_l$. Because $F_l$ is not a finite-horizon objective, there exists a stationary, optimal policy for it, considerably simplifying the learning problem.\footnote{This is the main reason why optimizing on an infinite-horizon objective, rather than a finite-horizon one, is an attractive choice.} Due to the difference between the learning and performance metrics, the policy that is optimal w.r.t. the learning metric does not need to be optimal w.r.t. the performance metric. We call the difference in performance between $\pi^*_l$ and $\pi^*$, as measured by $F$, the \emph{metric gap}:
$$\Delta_F = F(\pi^*, M) - F(\pi_l^*, M)$$
The relation between $\gamma$ and the metric gap will be analyzed in Section~\ref{sec:metric gap effect}.  

% Another useful concept to consider is how far an agent should `think ahead' in order to perform well on a specific task. We define the \emph{reasoning horizon} of a task to be the smallest horizon value an undiscounted, finite-horizon learning objective would need to have in order to do well on a task:
% $$h^r  = \min \{ h_l \,|   \Delta_F  < \epsilon\, ; F, F_l(\,\,\cdot\,\,; h_l)\}\,,$$
% where $\epsilon$ is some small value that sets the boundary for what is considered `doing well'. Besides the trivial case of a performance metric with a small horizon,
% the reasoning horizon can be small when for example there are repetitions in a task. The concept of the reasoning horizon generalizes the well-known concept of \emph{sparse-reward domains}, in the sense that any sparse-reward domain has a long reasoning horizon. 

We consider model-free, value-based methods. These are methods that aim to find a good policy by iteratively improving an estimate of the optimal action-value function $Q^*$, which, generally, predicts the expected discounted sum of rewards under the optimal policy $\pi^*_l$ conditioned on state-action pairs. The canonical example is Q-learning \citep{Watkins:1992Q}, which updates its estimates as follows:
\begin{equation}
Q_{t+1}(s_t, a_t) :=   (1-\alpha) Q_t(s_t, a_t) +
\alpha \left(r_{t} + \gamma \max_{a'} Q_t(s_{t+1},a')\right) \,,
\label{eq:ql_update}
\end{equation}
where $\alpha \in [0, 1]$ is the step-size. The action-value function is commonly estimated using a function approximator with weight vector $\theta$: $Q(s,a; \theta)$. Deep Q-Networks (DQN) \citep{mnih:nature15} use a deep neural network as function approximator and iteratively improve an estimate of  $Q^*$ by minimizing a sequence of loss functions:
\begin{align}
\mathcal{L}_i(\theta_i) &= \mathbb{E}_{s,a,r,s'} [(y^{DQN}_i  - Q(s,a; \theta_i))^2]\,, \label{eq:regular loss function} \\
\textnormal{with}\qquad y^{DQN}_i &= r + \gamma \max_{a'} Q(s',a' ; \theta_{i-1}),
\end{align}
The weight vector from the previous iteration, $\theta_{i-1}$, is encoded using a separate target network. 

\vspace{-\spaceSec em}
\section{Analysis of Discount Factor Effects}
\vspace{-\spaceSec em}

%We start by analyzing the effect of the discount factor on the metric gap. Then, we consider effects of the discount factor on optimizing the (discounted) learning objective.

\vspace{-\spaceSubSec em}
\subsection{Effect on Metric Gap}
\label{sec:metric gap effect}
\vspace{-\spaceSubSec em}

\begin{figure}[tbh]
\begin{center}
\includegraphics[height=2.6cm]{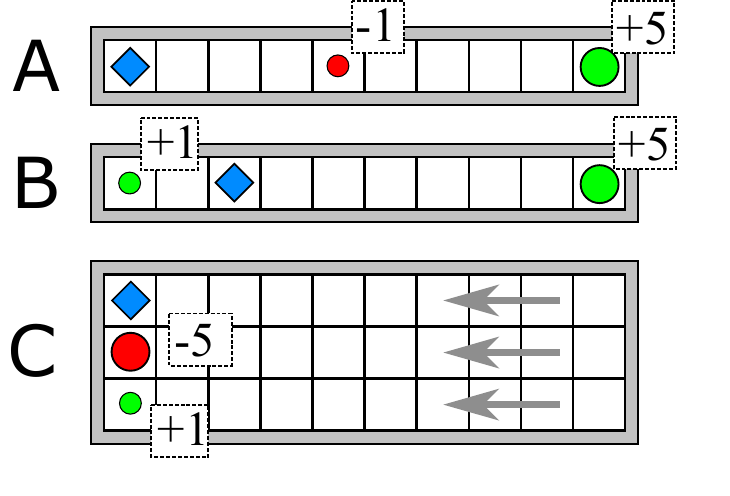}
\includegraphics[height=2.6cm]{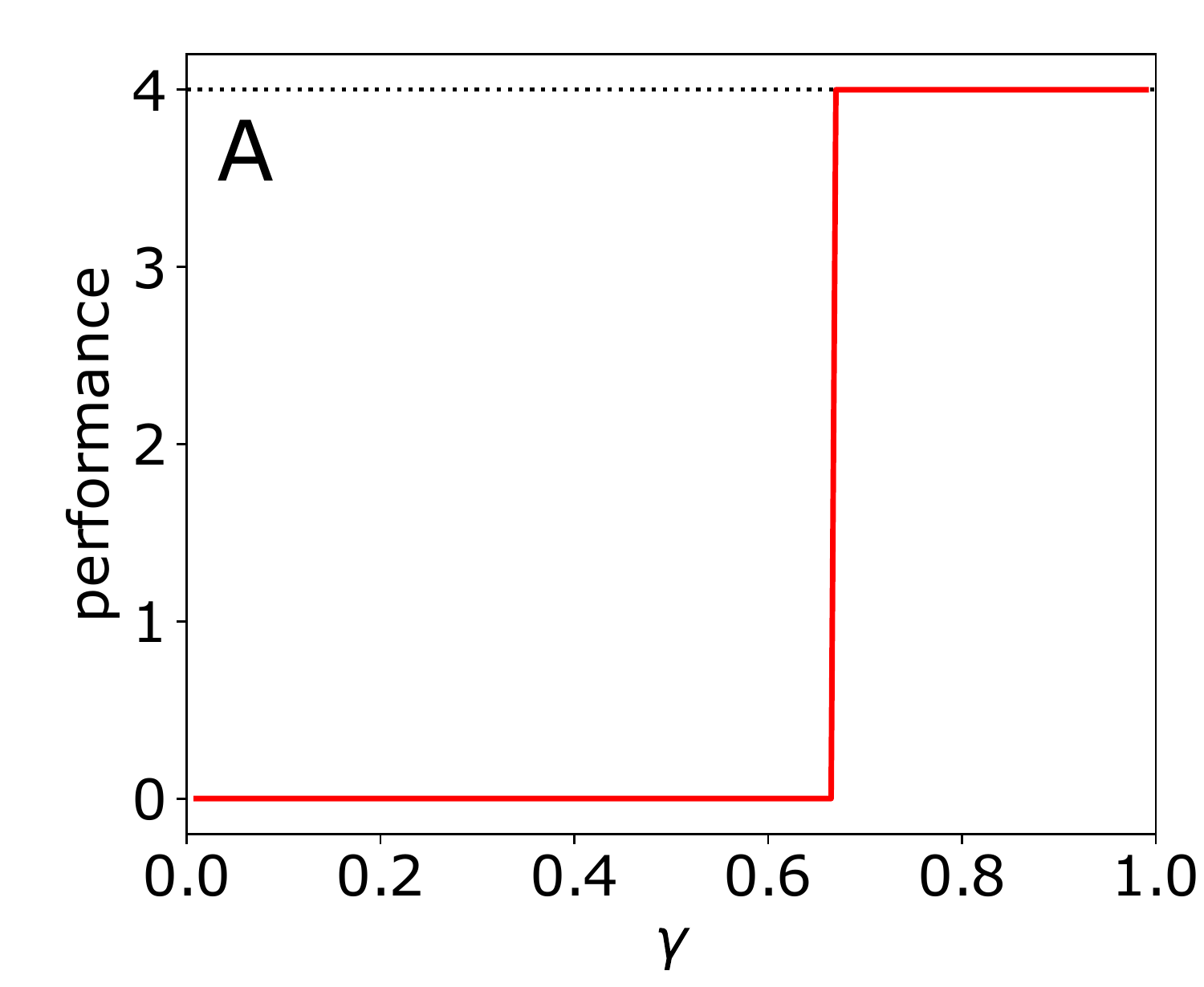}
\includegraphics[height=2.6cm]{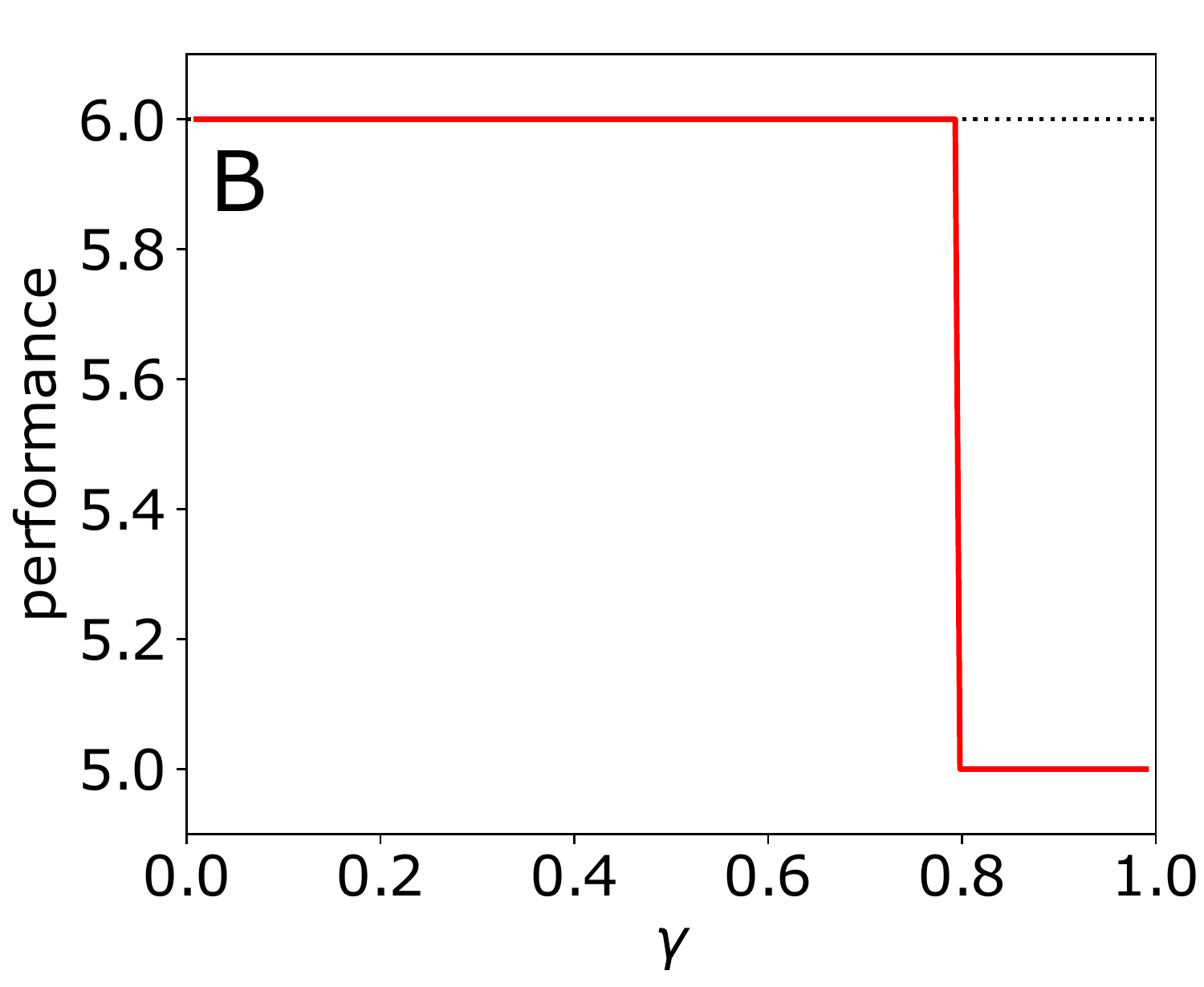}
\includegraphics[height=2.6cm]{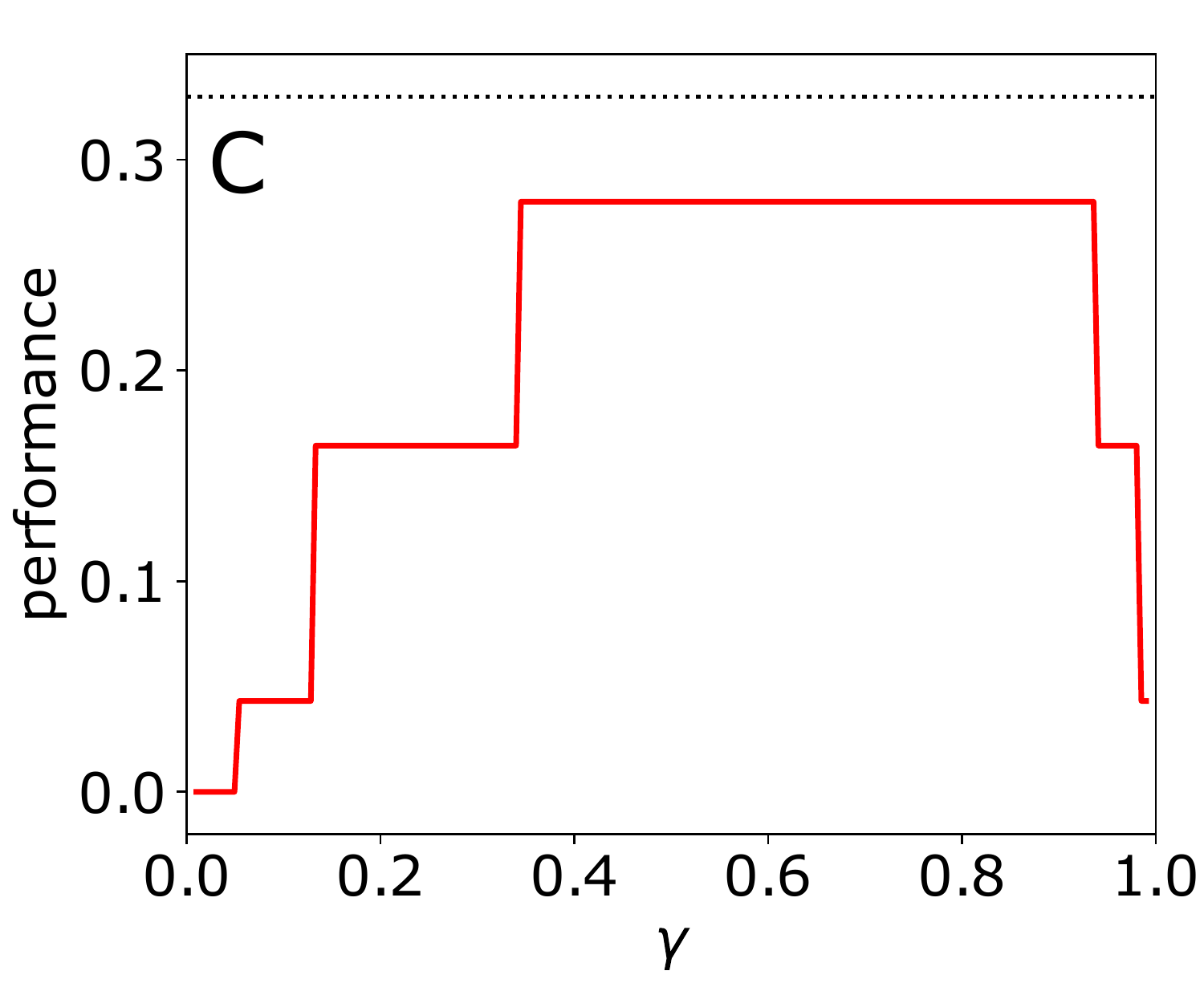}
\caption{Illustrations of three different tasks (blue diamond: starting position; green circle: positive object; red circle: negative object; gray arrows: wind direction; numbers indicate rewards). The graphs show the performance---as measured by $F$---on these tasks for $\pi^*$ (black, dotted line) and $\pi^*_l$ (red, solid line) as function of the discount factor of the learning metric. The difference between the two represents the metric gap.}
\label{fig:motivation tasks} 
\end{center}
\end{figure}

The question that is central to this section is the following: given a finite-horizon, undiscounted performance metric, what can be said about the relation between the discount factor of the learning metric and the metric gap? 

To study this problem, we designed a variety of different tasks and measured the dependence between the metric gap and the discount factor. In Figure \ref{fig:motivation tasks}, we illustrate three of those tasks, as well as the metric gap on those tasks as function of the discount factor. In each task, an agent, starting from a particular position,  has to collect rewards by collecting the positive objects while avoiding the negative objects. The transition dynamics of tasks A and B is deterministic; whereas, in task C wind blows in the direction of the arrows, making the agent move towards left with a 40\% chance, regardless of its performed action. For all three tasks, the horizon of the performance metric is 12. 

On task A, where a small negative reward has to be traded off for a large positive reward that is received later, high discount factors result in a smaller metric gap. By contrast, on task B, low discount factors result in a smaller metric gap. The reason is that for high discount factors the optimal learning policy takes the longer route by first trying to collect the large object, before going to the small object. However, with a performance metric horizon of 12, there is not enough time to take the long route and get both rewards. The low discount factor takes a shorter route by first going to the smaller object and is able to collect all objects in time. On task C, a trade-off has to be made between the risk of falling into the negative object (due to domain stochasticity) versus taking a longer detour that minimizes this risk. On this task, the optimal policy $\pi^*$ is non-stationary (the optimal action depends on the time step). However, because the learning objective $F_l$ is not finite-horizon, it has a stationary optimal policy $\pi_l^*$. Hence, the metric gap cannot be reduced to 0 for any value of the discount factor. The best discount factor is something that is not too high nor too low.

While the policy $\pi_l^*$ is derived from an infinite-horizon metric, this does not preclude it from being learned with finite-length training episodes. As an example, consider using Q-learning to learn $\pi_l^*$ for any of the tasks from Figure \ref{fig:motivation tasks}. With a uniformly random behavior policy and training episodes of length 12 (the same as the horizon of the performance metric), there is a non-zero probability for each state-action pair that it will be visited within an episode. Hence, with the right step-size decay schedule, convergence in the limit can be guaranteed \citep{jaakkola1994convergence}. A key detail to enable this is that the state that is reached at the final time step is not treated as a terminal state (which has value 0 by default), but normal bootstrapping occurs \citep{pmlr-v80-pardo18a}.

A finite-horizon performance metric is not essential to observe strong dependence of the metric gap on $\gamma$. For example, if on task B the performance metric would measure the number of steps it takes to collect all objects, a similar graph is obtained. In general, the examples in this section demonstrate that the best discount factor is task-dependent and can be anywhere in the range between $0$ and $1$.

% In general, an agent with a high discount factor simply makes different trade-offs than one with a low discount factor, which can result in better or worse performance, depending on what performance metric is used. 

% The examples above demonstrate that the best discount factor is task-dependent and can be anywhere in the range between $0$ and $1$.

\vspace{-\spaceSubSec em}
\subsection{Optimization Effects}
\label{sec:optimization}
\vspace{-\spaceSubSec em}

\begin{figure}[t]
\begin{center}
\includegraphics[width=7.5cm]{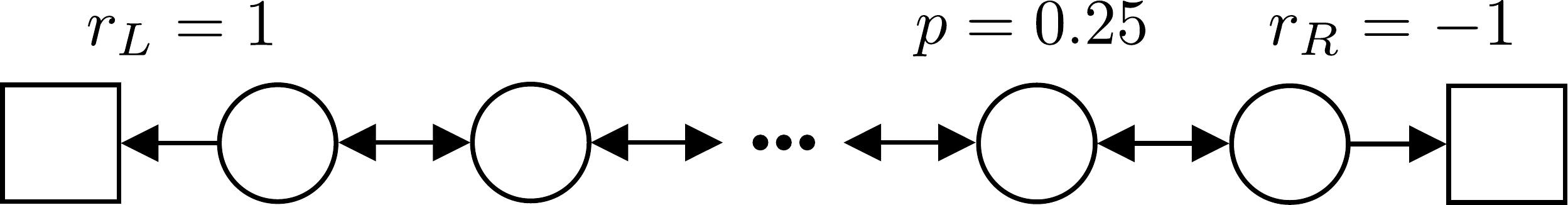}
\caption{Chain task consisting of 50 states and two terminal ones. Each (non-terminal) state has two actions: $a_L$ which results in transitioning to the left with probability $1-p$ and to the right with probability $p$, and vice versa for the other action, $a_R$. All rewards are 0, except for transitioning to the far-left or far-right terminal states that result in $r_L$ and $r_R$, respectively.}
\label{fig:main toy task} 
\end{center}
\vspace{-15pt}
\end{figure}

The performance of $\pi_l^*$ gives the theoretical limit of what the agent can achieve given its learning metric. However, the discount factor also affects the optimization process; for some discount factors, finding $\pi_l^*$ could be more challenging than for others. In this section, using the task shown in Figure~\ref{fig:main toy task}, we evaluate the correlation between the discount factor and how hard it could be to find $\pi_l^*$. It is easy to see that the policy that always takes the left action $a_L$ maximizes 
%the (discounted) 
both discounted and undiscounted
sum of rewards for any discount factor or horizon value, respectively. We define the learning metric $F_l$ as before (\ref{eq:metrics}), but use a different performance metric $F$. Specifically, we define $F$ to be $1$ if the policy takes $a_L$ in every state, and $0$ otherwise. The metric gap for this setting of $F$ and $F_l$ is $0$, with the optimal performance (for $\pi^*$ and $\pi^*_l$) being $1$.

\begin{wrapfigure}{}{0.32\textwidth}
\vspace{-15pt}
\centering
\includegraphics[width=4.5cm]{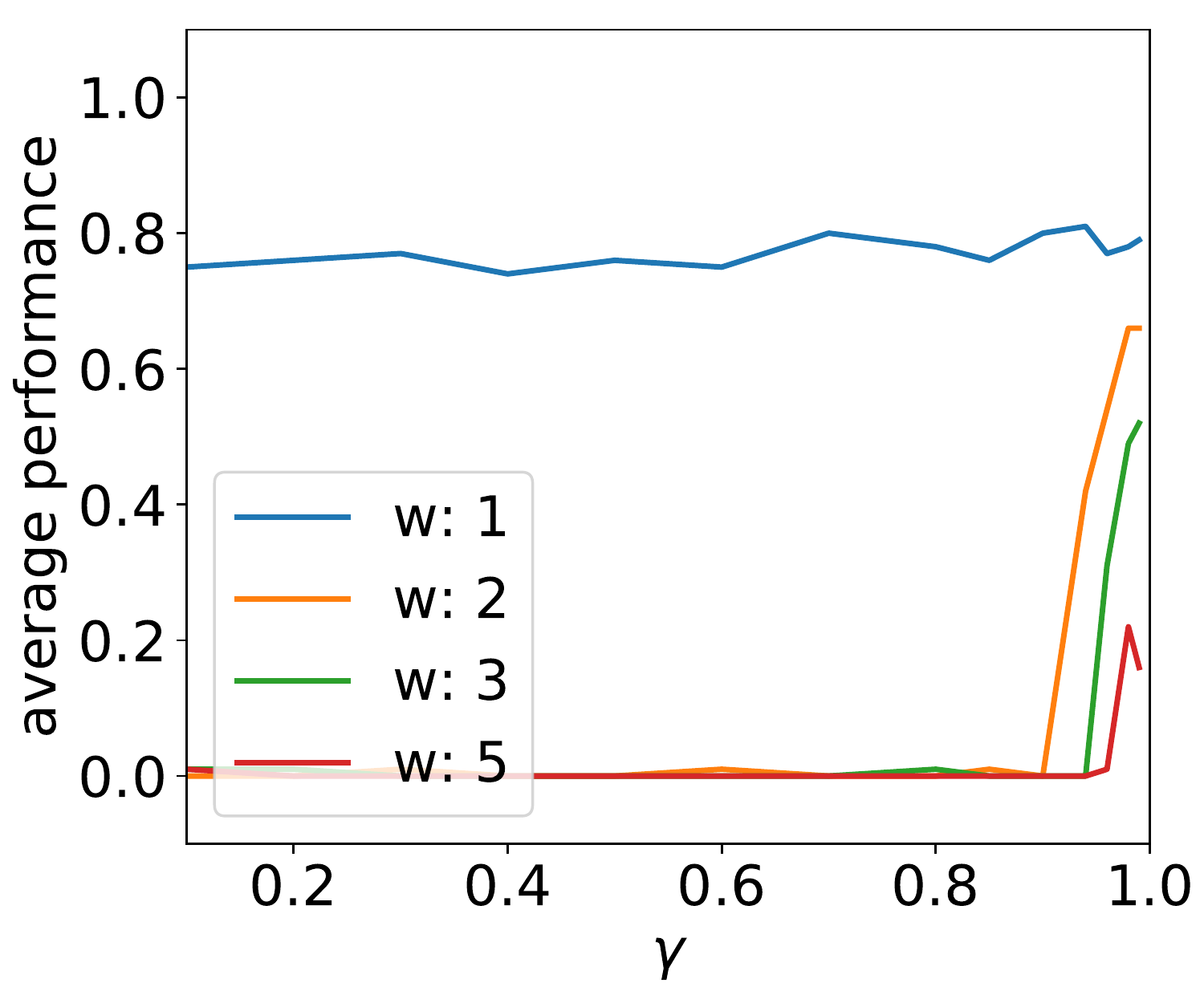}
\includegraphics[width=4.5cm]{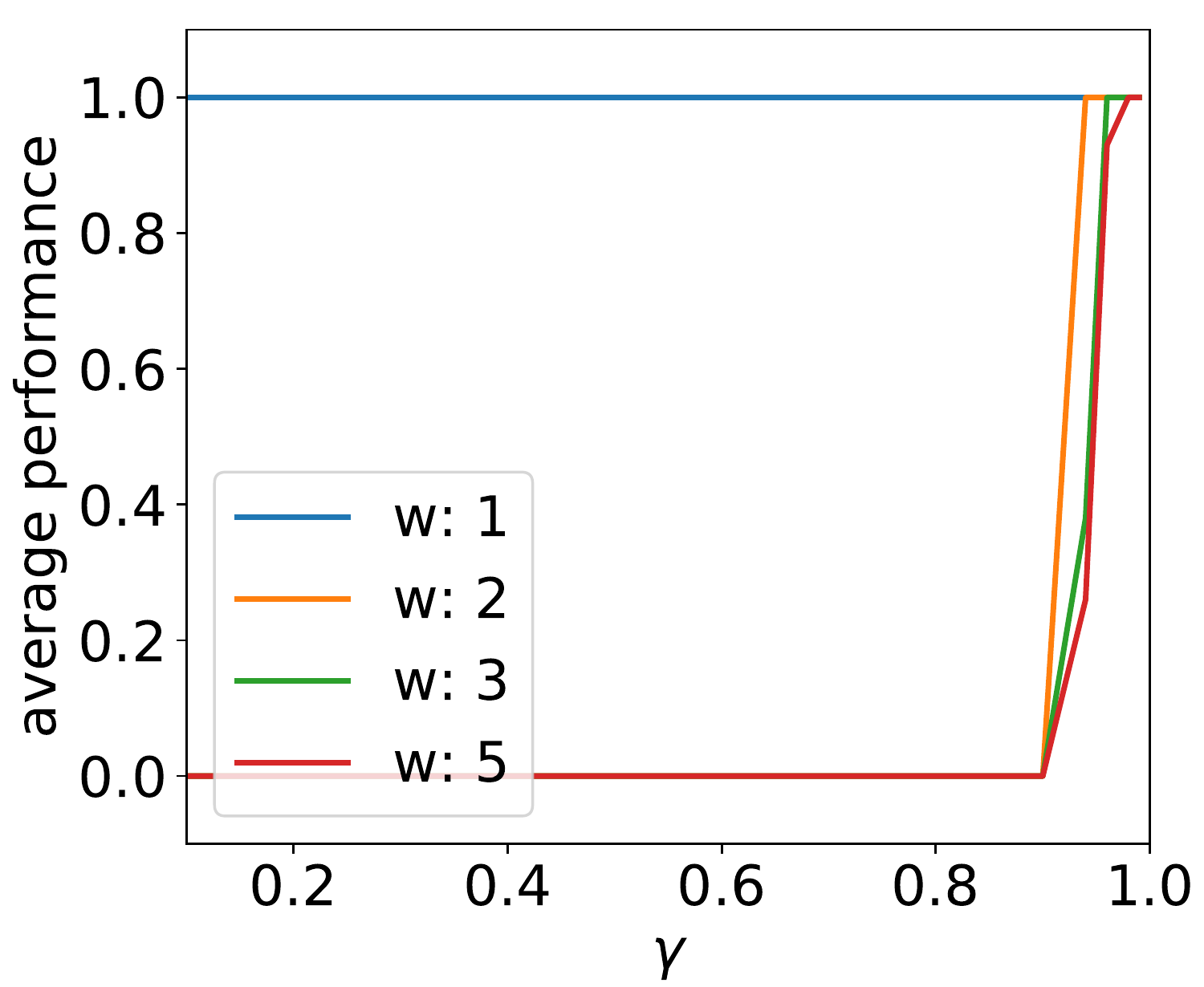}
\caption{Early performance (top) and final performance (bottom) on the chain task.}
\label{fig:reg result}
\vspace{-15pt}
\end{wrapfigure}

To study the optimization effects under function approximation, we use linear function approximation with features constructed by tile-coding \citep{sutton1996generalization}, using tile-widths of $1$, $2$, $3$, and $5$. A tile-width of $w$ corresponds to a binary feature that is non-zero for $w$ neighbouring states and zero for the remaining ones. The number and offset of the tilings are such that any value function can be represented. Hence, error-free reconstruction of the optimal action-value function is possible in principle, for any discount factor. Note that for a width of $1$, the representation reduces to a tabular one.

To keep the experiment as simple as possible, we remove exploration effects by performing update sweeps over the entire state-action space (using a step-size of $0.001$) and measure performance at the end of each update sweep. Figure~\ref{fig:reg result} shows the performance during early learning (average performance over the first $10,000$ sweeps) as well as the final performance (average between sweeps $100,000$ and $110,000$).

These experiments demonstrate a common empirical observation: when using function approximation, low discount factors do not work well in sparse-reward domains. More specifically, the main observations are: 1) there is a sharp drop in final performance for discount factors below some threshold; 2) this threshold value depends on the tile-width, with larger ones resulting in worse (i.e., higher) threshold values; and 3) the tabular representation performs well for all discount factors.

It is commonly believed that the \emph{action gap} has a strong influence on the optimization process \citep{bellemare2016:actiongap,farahmand2011:actiongap}. The action gap of a state $s$ is defined as the difference in $Q^*$ between the best and the second best actions at that state.
To examine this common belief, we start by evaluating two straightforward hypotheses involving the action gap: 1) lower discount factors cause poor performance because they result in smaller action gaps; 2) lower discount factors cause poor performance because they result in smaller relative action gaps (i.e, the action gap of a state divided by the maximum action-value of that state). Since both hypotheses are supported by the results from Figure~\ref{fig:reg result}, we performed more experiments to test them. To test the first hypothesis, we performed the same experiment as above, but with rewards that are a factor $100$ larger. This in turn increases the action gaps by a factor $100$ as well. Hence, to validate the first hypothesis, this change should improve (i.e., lower) the threshold value where the performance falls flat.  To test the second hypothesis, we pushed all action-values up by $100$ through additional rewards, reducing the relative action-gap. Hence, to validate the second hypothesis, performance should degrade for this variation. However, neither of the modifications caused significant changes to the early or final performance, invalidating these hypotheses. The corresponding graphs can be found in Appendix B.

\begin{wrapfigure}{r}{0.32\textwidth}
\vspace{-10pt}
\centering
\includegraphics[width=4.5cm]{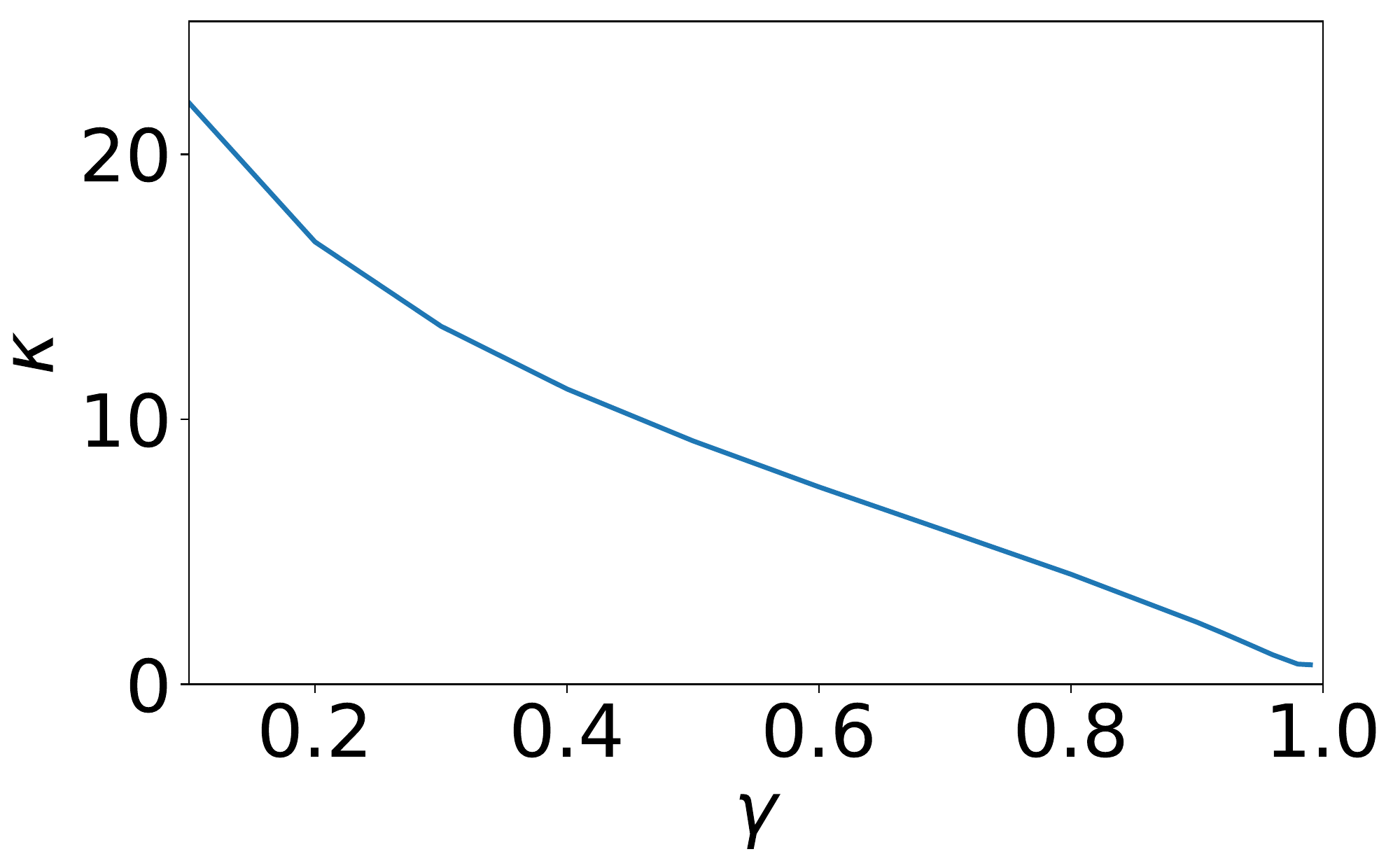}
\caption{Action-gap deviation as function of discount factor.}
\label{fig:kappa reg full}
\vspace{0pt}
\end{wrapfigure}

Because our two na\"ive action-gap hypotheses have failed, we propose an alternative hypothesis: lower discount factors cause poor performance because they result in a larger difference in the action-gap sizes across the state-space. To illustrate the statement about the difference in action-gap sizes, we define a metric, which we call the \emph{action-gap deviation} $\kappa$, that aims to capture the notion of action-gap variations. Specifically, let $X$ be a random variable and let $\mathcal{S}^+ \subseteq \mathcal{S}$  be the subset of states that have a non-zero action gap.  $X$ draws uniformly at random a state $s \in \mathcal{S}^+$ and outputs $\log_{10}\left(AG(s)\right)$, where $AG(s)$ is the action gap of state $s$. We now define $\kappa$ to be the standard deviation of the variable $X$. Figure~\ref{fig:kappa reg full} plots $\kappa$ as function of the discount factor for the task in Figure~\ref{fig:main toy task}.

To test this new hypothesis, we have to develop a method that reduces the action-gap deviation $\kappa$ for low discount factors, without changing the optimal policy. %This will be done in the next section.
We do so in the next section.

\vspace{-\spaceSec em}
\section{Logarithmic Q-learning}
\vspace{-\spaceSec em}
\label{sec:q-learning}

In this section, we introduce our new method, logarithmic Q-learning, which reduces the action-gap deviation $\kappa$ for sparse-reward domains. We present the method in three steps, in each step adding a layer of complexity
%while extending 
in order to extend 
the generality of the method. In Appendix A, we prove convergence of the method in its most general form. 
%First, 
As the first step, we now consider domains with deterministic dynamics and rewards that are either positive or zero.

\vspace{-\spaceSubSec em}
\subsection{Deterministic Domains with Positive Rewards}
\vspace{-\spaceSubSec em}

Our method is based on the same general approach as used by \cite{pohlen2018observe}: mapping the update target to a different space and performing updates in that space instead. We indicate the mapping function by $f$, and its inverse by $f^{-1}$. Values in the mapping space are updated as follows:
\begin{equation}
\tQ_{t+1}(s_t, a_t) :=   (1-\alpha) \tQ_t(s_t, a_t) +
\alpha f\left(r_{t} + \gamma \max_{a'}  f^{-1}\left( \tQ_t(s_{t+1},a')\right)\right) \,.
\label{eq:log_update}
\end{equation}
Note that $\tQ$ in this equation is not an estimate of an expected return; it is an estimate of an expected return mapped to a different space. To obtain a regular Q-value the inverse mapping has to be applied to $\tQ$. 
Because the updates occur in the mapping space, $\kappa$ is now measured w.r.t. $\tQ$. That is, the action gap of state $s$ is now defined in the mapping space as $\tQ(s,a_{best}) - \tQ(s,a_{2nd\,best})$.%\footnote{ our mapping function is a strictly increasing function, the best action in the regular space is also the best action in the mapping space.}

\begin{wrapfigure}{r}{0.32\textwidth}
\vspace{-10pt}
\centering
\includegraphics[width=4.5cm]{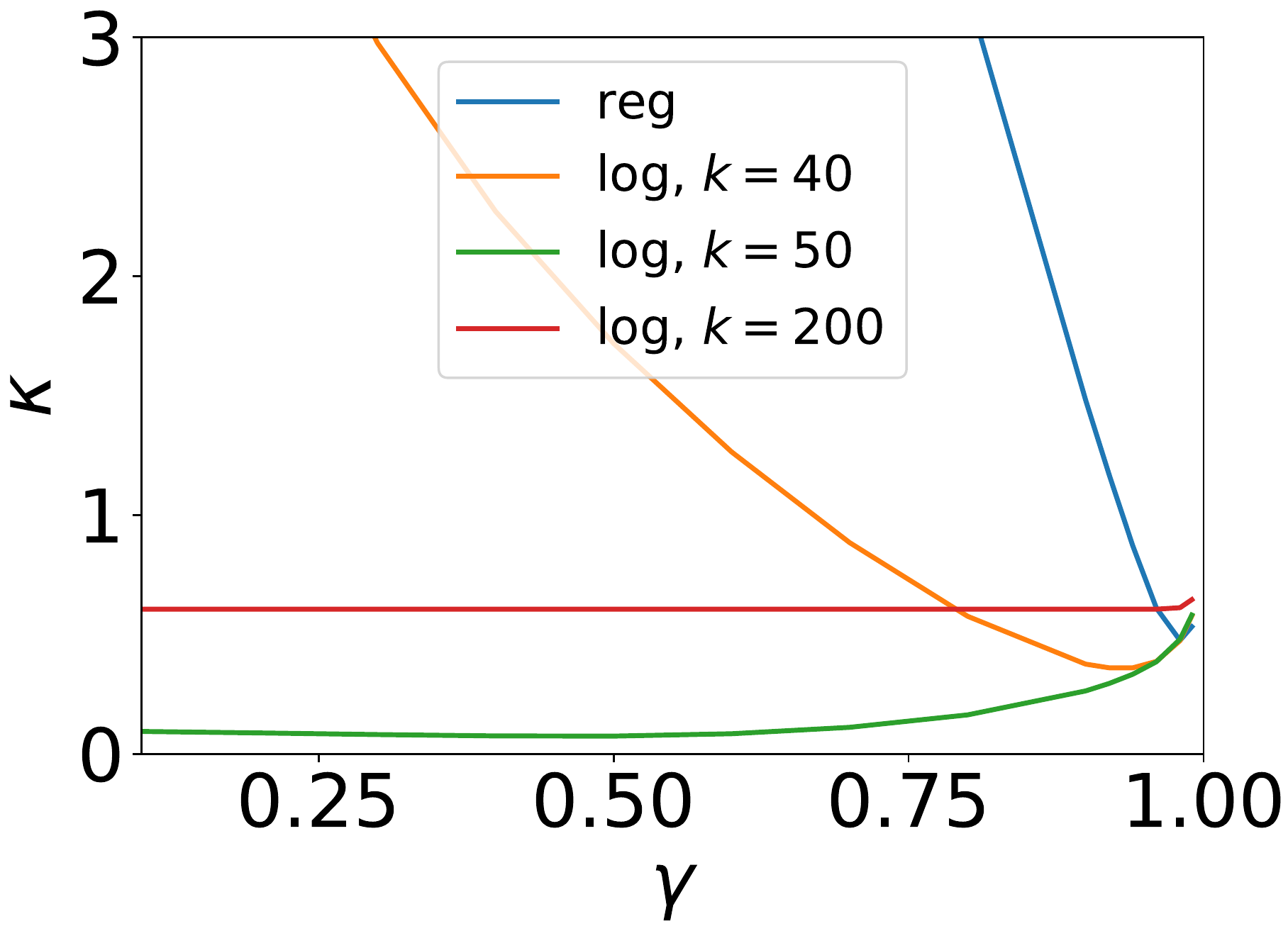}
\caption{Action-gap deviation as function of discount factor.}
\label{fig:kappa all det}
\vspace{-5pt}
\end{wrapfigure}

To reduce $\kappa$, we propose to use a logarithmic mapping function. Specifically, we propose the following mapping function:
\begin{equation}
f(x) := c\,\ln(x + \gamma^k) + d\,,
\label{eq:mapping}
\end{equation}
with inverse function: $f^{-1}(x) = e^{(x - d)/c} - \gamma^k\,$, where $c$, $d$, and $k$ are mapping hyper-parameters.

% In the supplementary material, we prove that $\tQ$ converges to $f(Q^*)$ under standard conditions when mapping function (\ref{eq:mapping}) is used with update rule  (\ref{eq:log_update}) and the environment is deterministic. If the environment is stochastic $\tQ$ convergences to an underestimate of $f(Q^*)$ due to Jensen's inequality. Another restriction of the mapping function that $x$ should be larger than  $\gamma^k$ to avoid the input of the $\ln(\cdot)$ function to be negative. Hence, we only consider domains with rewards $\geq 0$ in this section.

To understand the effect of (\ref{eq:mapping}) on $\kappa$, we plot $\kappa$, based on action gaps in the logarithmic space, on a variation of the chain task (Figure~\ref{fig:main toy task})  that uses $r_R = 0$ and $p = 0$. We also plot $\kappa$ based on actions in the regular space. Figure~\ref{fig:kappa all det} shows that with an appropriate value of $k$, the action-gap deviation can almost be reduced to $0$ for low values of $\gamma$. Setting $k$ too high increases the deviation a little, while setting it too low increases it a lot for low discount factors. In short, $k$ controls the smallest Q-value that can still be accurately represented (i.e., for which the action gap in the log-space is still significant). Roughly, the smallest value that can still be accurately represented is about $\gamma^k$. In other words, the cut-off point lies approximately at a state from which it takes $k$ time steps to experience a $+1$ reward. Setting $k$ too high causes actions that have $0$ value in the regular space to have a large negative value in the log-space. This can increase the action gap substantially for the corresponding states, thus, resulting in an overall increase of the action-gap deviation.

The parameters $c$ and $d$ scale and shift values in the logarithmic space and do not have any effect on the action-gap deviation. The parameter $d$ controls the initialization of the Q-values. Setting $d$ as follows:
\begin{equation}
d = -c\,\ln(q_{init} + \gamma^k)\,,
\label{eq:d setting}
\end{equation}
ensures that $f^{-1}(0) = q_{init}$ for any value of $c, k$, and $\gamma$. 
This can be useful in practice, e.g., when using neural networks to represent $\tQ$, as it enables standard initialization methods (which produce output values around $0$) while still ensuring that the initialized $\tQ$ values correspond with $q_{init}$ in the regular space. The parameter $c$ scales values in the log-space. For most tabular and linear methods, scaling values does not affect the optimization process. Nevertheless, in deep RL methods more advanced optimization techniques are commonly used and, thus, such scaling can impact the optimization process significantly. In all our experiments, except the deep RL experiments, we fixed $d$ according to the equation above with $q_{init} = 0$ and used $c=1$.

In stochastic environments, the approach described in this section causes issues, because averaging over stochastic samples in the log-space produces an underestimate compared to averaging in the regular space and then mapping the result to the log-space. Specifically, if $X$ is a random variable, $\mathbb{E}\left[ \ln(X)\right] \leq  \ln\left(\mathbb{E}[X]\right)$ (i.e., Jensen's inequality). Fortunately, within our specific context, there is a way around this limitation that we discuss in the next section.

\vspace{-\spaceSubSec em}
\subsection{Stochastic Domains with Positive Rewards}
\vspace{-\spaceSubSec em}

The step-size $\alpha$ generally conflates two forms of averaging: averaging of stochastic update targets due to environment stochasticity,  and, in the case of function approximation, averaging over different states. To amend our method for stochastic environments, ideally, we would separate these forms of averaging and perform the averaging over stochastic update targets in the regular space and the averaging over different states in the log-space. While such a separation is hard to achieve, the approach presented below, which is inspired by the above observation, achieves many of the same benefits. In particular, it enables convergence of $\widetilde Q$ to $f(Q^*)$, even when the environment is stochastic.

Let $\beta_{log}$ be the step-size for averaging in the log-space, and $\beta_{reg}$ be the step-size for averaging in the regular space. We amend the approach from the previous section by computing an alternative update target that is based on performing an averaging operation in the regular space. Specifically, the update target $U_t$ is transformed into an alternative update target $\hat U_t$ as follows:
\begin{equation}
\hat U_t :=  f^{-1}(\tQ_t(s_t, a_t)) + \beta_{reg} \left(U_t - f^{-1}(\tQ_t(s_t,a_t) \right)\,,
\label{eq:beta_reg_update}
\end{equation}
with $U_t :=  r_t +  \gamma \max_{a'}  f^{-1}( \tQ_t(s_{t+1},a'))$. The modified update target $\hat U_t$ is used for the update in the log-space:
\begin{equation}
\tQ_{t+1}(s_t, a_t) :=   \tQ_t(s_t, a_t) + \beta_{log} \left(f(\hat U_t) - \tQ_t(s_t, a_t)\right) \,.
\label{eq:beta_log_update}
\end{equation}
Note that if $\beta_{reg}=1$, then $\hat U_t = U_t$, and update (\ref{eq:beta_log_update}) reduces to update (\ref{eq:log_update}) from the previous section, with $\alpha = \beta_{log}$.

\begin{wrapfigure}{r}{0.32\textwidth}
\vspace{-15pt}
\centering
\includegraphics[width=4cm]{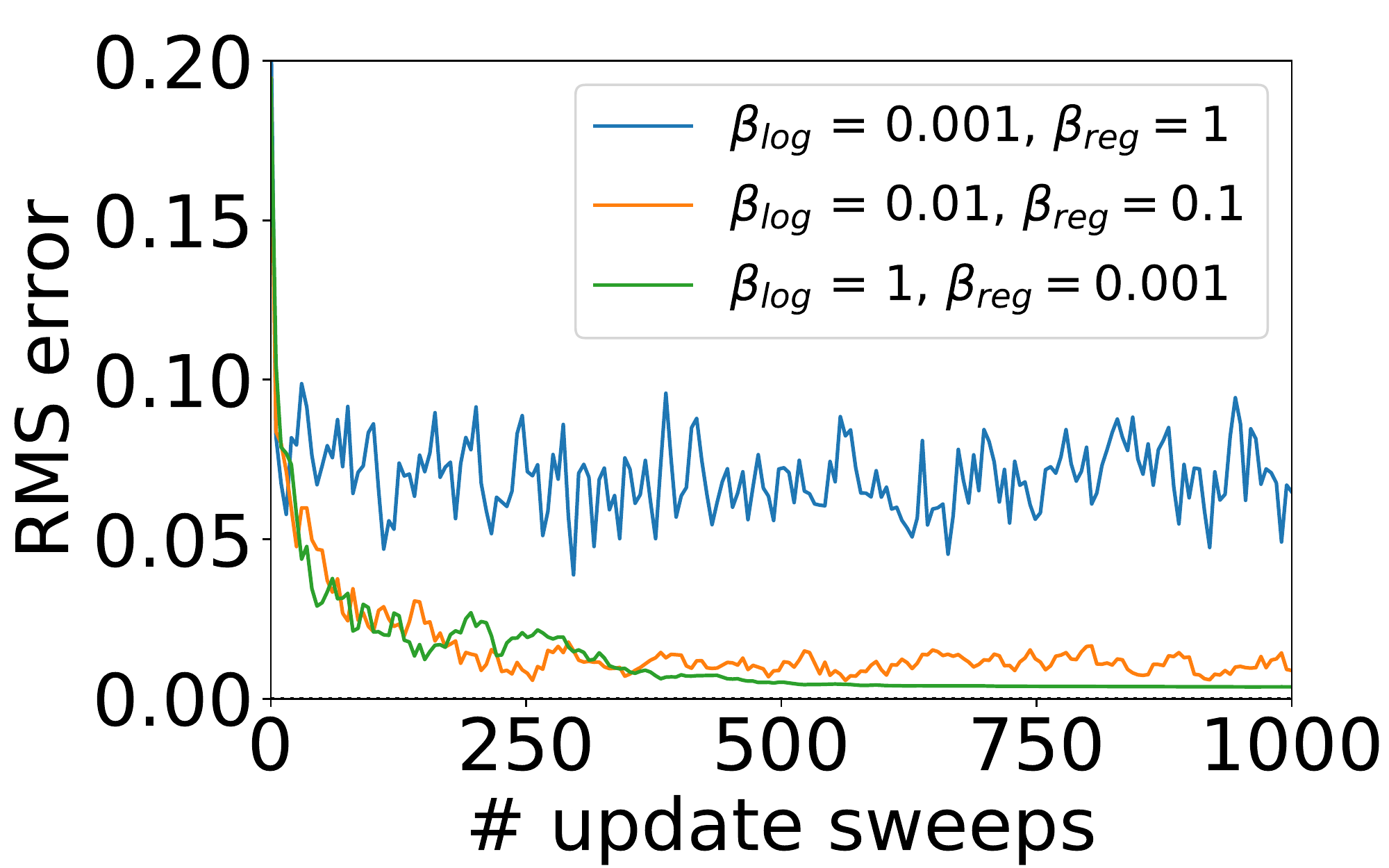}
\vspace{-5pt}
\caption{RMS error.}
\label{fig:rms error}
\vspace{-10pt}
\end{wrapfigure}

The conditions for convergence are discussed in the next section, but one of the conditions is that $\beta_{reg}$ should go to $0$ in the limit. From a more practical point of view, when using fixed values for $\beta_{reg}$ and $\beta_{log}$, $\beta_{reg}$ should be set sufficiently small to keep underestimation of values due to the averaging in the log-space under control. To illustrate this, we plot the RMS error on a positive-reward variant of the chain task ($r_R=0, r_L = +1, p = 0.25$). The RMS error plotted is based on the difference between $f^{-1}(\tQ(s,a))$ and $Q^*(s,a)$ over all state-action pairs. We used a tile-width of 1, corresponding with a tabular representation, and used $k=200$. Note that for $\beta_{reg} = 1$, which reduces the method to the one from the previous section, the error never comes close to zero.

\vspace{-\spaceSubSec em}
\subsection{Stochastic Domains with Positive and/or Negative Rewards}
\vspace{-\spaceSubSec em}

We now consider the general case where the rewards can be both positive or negative (or zero). It might seem that we can generalize to negative rewards simply by replacing $x$ in the mapping function (\ref{eq:mapping}) by $x + D$, where $D$ is a sufficiently large constant that prevents $x +D$ from becoming negative. The problem with this approach, as we will demonstrate empirically, is that it does not decrease $\kappa$ for low discount factors. Hence, in this section we present an alternative approach, based on decomposing the Q-value function into two functions.

Consider a decomposition of the reward $r_t$ into two components, $r^+_t$ and $r^-_t$, as follows:
\begin{equation}
r^+_t:=
\begin{cases}
r_t & \mbox{if } r_t \geq 0\\
0  & \mbox{otherwise }\ \\
\end{cases}
\quad ; \quad
r^-_t :=
\begin{cases}
|r_t| & \mbox{if } r_t < 0\\
0  & \mbox{otherwise }\ \\
\end{cases}
\,.
\label{eq:reward decomposition}
\end{equation}
Note that $r^+_t$ and $r^-_t$ are always non-negative and that $r_t = r^+_t - r^-_t$ at all times. By decomposing the observed reward in this manner, these two reward components can be used to train two separate Q-value functions: $\tQ^+$, which represents the value function in the mapping space corresponding to $r_t^+$, and $\tQ^-$, which plays the same role for $r_t^-$. To train these value functions, we construct the following update targets:
\begin{equation}
U_t^+ :=  r_t^+ +  \gamma  f^{-1}\left( \tQ^+_t(s_{t+1},\tilde a_{t+1})\right)\quad ; \quad
U_t^- :=  r_t^- +  \gamma  f^{-1}\left( \tQ^-_t(s_{t+1},\tilde a_{t+1})\right)
\label{eq:plus_min_U}
\end{equation}
with $\,\,\tilde a_{t+1} := \argmax_{a'} \left( f^{-1}(\tQ_{t}^+(s_{t+1},a')) - f^{-1}(\tQ_{t}^-(s_{t+1},a'))\right)$. These update targets are modified into $\hat U_t^+$ and $\hat U_t^-$, respectively, based on (\ref{eq:beta_reg_update}), which are then used to update $\tQ^+$ and $\tQ^-$, respectively, based on (\ref{eq:beta_log_update}). Action-selection at time $t$ is based on $Q_t$, which we define as follows:
\begin{equation}
Q_t(s,a) := f^{-1}\left(\tQ_{t}^+(s,a)\right) - f^{-1}\left(\tQ_{t}^-(s,a)\right)
\label{eq:Q definition}
\end{equation}

In Appendix A, we prove convergence of logarithmic Q-learning under similar conditions as regular Q-learning. In particular, the product $\beta_{log,t}\cdot \beta_{reg,t}$ has to satisfy the same conditions as $\alpha_t$ does for regular Q-learning. There is one additional condition on $\beta_{reg,t}$, which states that it should go to zero in the limit.

% The following theorem states the conditions of convergence for our method. The proof of the theorem can be found in the supplementary material.
% \begin{theorem}
% If the reward $r_t$ is decomposed according to (\ref{eq:reward decomposition}), and the functions  $\tQ^+$ and $\tQ^-$ each are updated based on (\ref{eq:plus_min_U}), (\ref{eq:beta_reg_update}), and (\ref{eq:beta_log_update}), then $Q_t$, as defined by (\ref{eq:Q definition}) converges to $Q^*$ w.p. 1 if the following conditions hold: 
% \begin{enumerate}
%     \item $0\leq \beta_{log,t} \cdot \beta_{reg,t} \leq 1$
%     \item $\sum_{t=0}^\infty  \beta_{log,t}\cdot \beta_{reg,t} = \infty$
%     \item $\sum_{t=0}^\infty (\beta_{log,t}\cdot \beta_{reg,t})^2 < \infty$
%     \item $\lim_{t\rightarrow\infty} \beta_{reg,t} = 0$
% \end{enumerate}
% % 1) $\,\, 0\leq \beta_{log,t} ; \beta_{reg,t} \leq 1\quad$; 2) $\,\,\sum_{t=0}^\infty  \beta_{log,t}\cdot \beta_{reg,t} = \infty\quad$; 3) $\,\,\sum_{t=0}^\infty (\beta_{log,t}\cdot \beta_{reg,t})^2 < \infty\quad$; 4) $\,\,\lim_{t\rightarrow\infty} \beta_{reg,t} = 0$.
% \label{eq:theorem}
% \end{theorem}

\begin{wrapfigure}{r}{0.32\textwidth}
\vspace{-12pt}
\centering
\includegraphics[width=4.5cm]{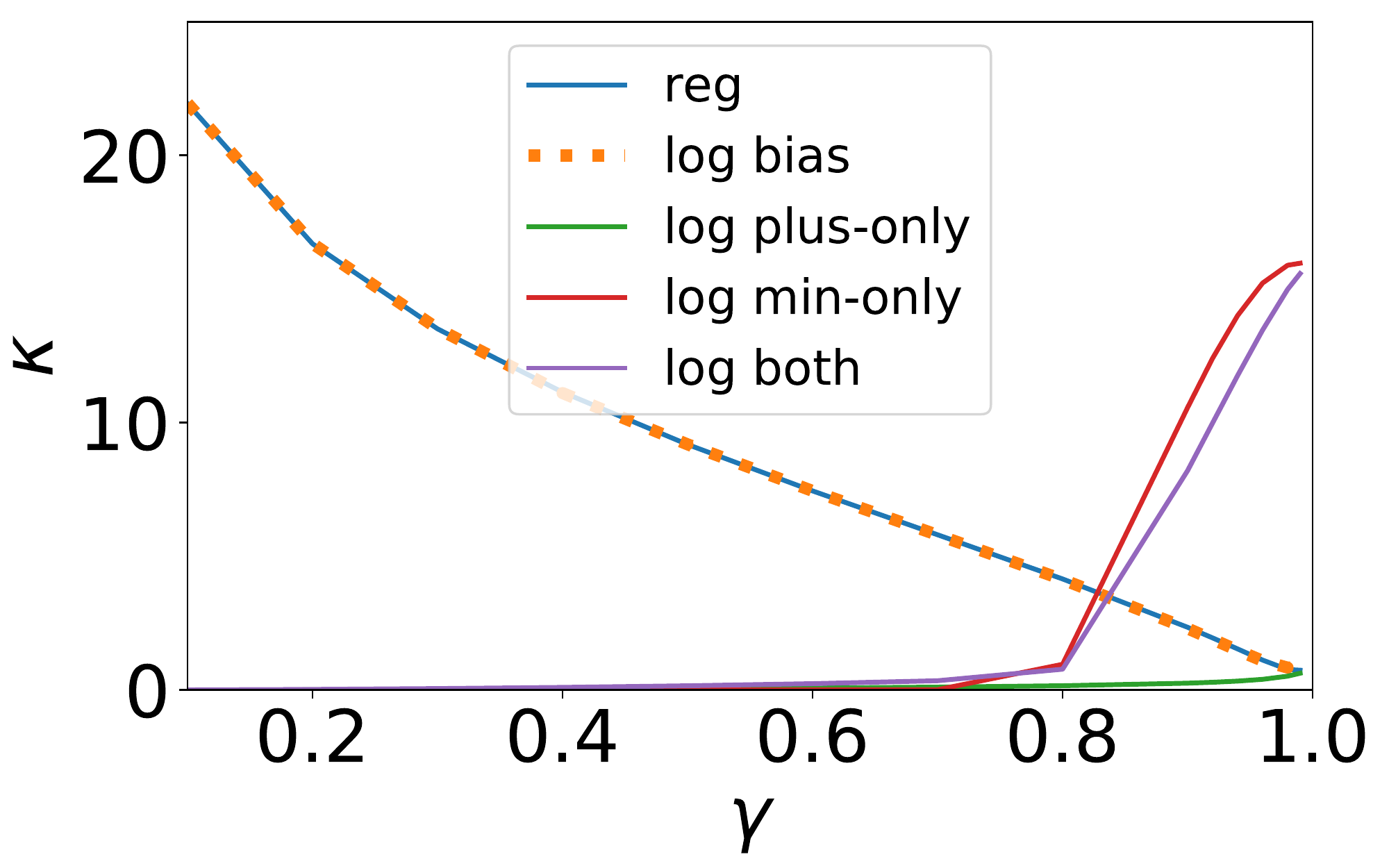}
\caption{Action-gap deviation as function of discount factor.}
\label{fig:kappa dqn full}
\vspace{0pt}
\end{wrapfigure}

We now compute $\kappa$ for the full version of the chain task. Because there are two functions, $\tQ^+$ and $\tQ^-$, we have to generalize the definition of $\kappa$ to this situation. We consider three generalizations: 1) $\kappa$ is based on the action-gaps of $\tQ^+$ (`log plus-only'); 2) $\kappa$ is based on the action-gaps of $\tQ^-$ (`log min-only'); and 3) $\kappa$ is based on the action-gaps of both $\tQ^-$ and $\tQ^+$ (`log both'). Furthermore, we plot a version that resolves the issue of negative rewards na\"ively, by adding a value $D=1$ to the input of the log-function (`log bias'). We plot $\kappa$ for these variants in Figure~\ref{fig:kappa dqn full}, using $k=200$, together with $\kappa$ for regular Q-learning (`reg'). Interestingly, only for the `log plus-only' variant $\kappa$ is small for all discount factors. Further analysis showed that the reason for this is that under the optimal policy, the chance that the agent moves from a state close to the positive terminal state to the negative terminal state is very small, which means that $k=200$ is too small to make the action-gaps for $\tQ^-$ homogeneous. However, as we will see in the next section, the performance with $k=200$ is good for all discount factors, demonstrating that not having homogeneous action-gaps for $\tQ^-$ is not a huge issue. We argue that this could be because of the behavior related to the nature of positive and negative rewards: it might be worthwhile to travel a long distance to get a positive reward, but avoiding a negative reward is typically a short-horizon challenge.

\newpage
\vspace{-\spaceSec em}
\section{Experiments}
\vspace{-\spaceSec em}

\begin{wrapfigure}{}{0.32\textwidth}
\vspace{-15pt}
\centering
\includegraphics[width=4.5cm]{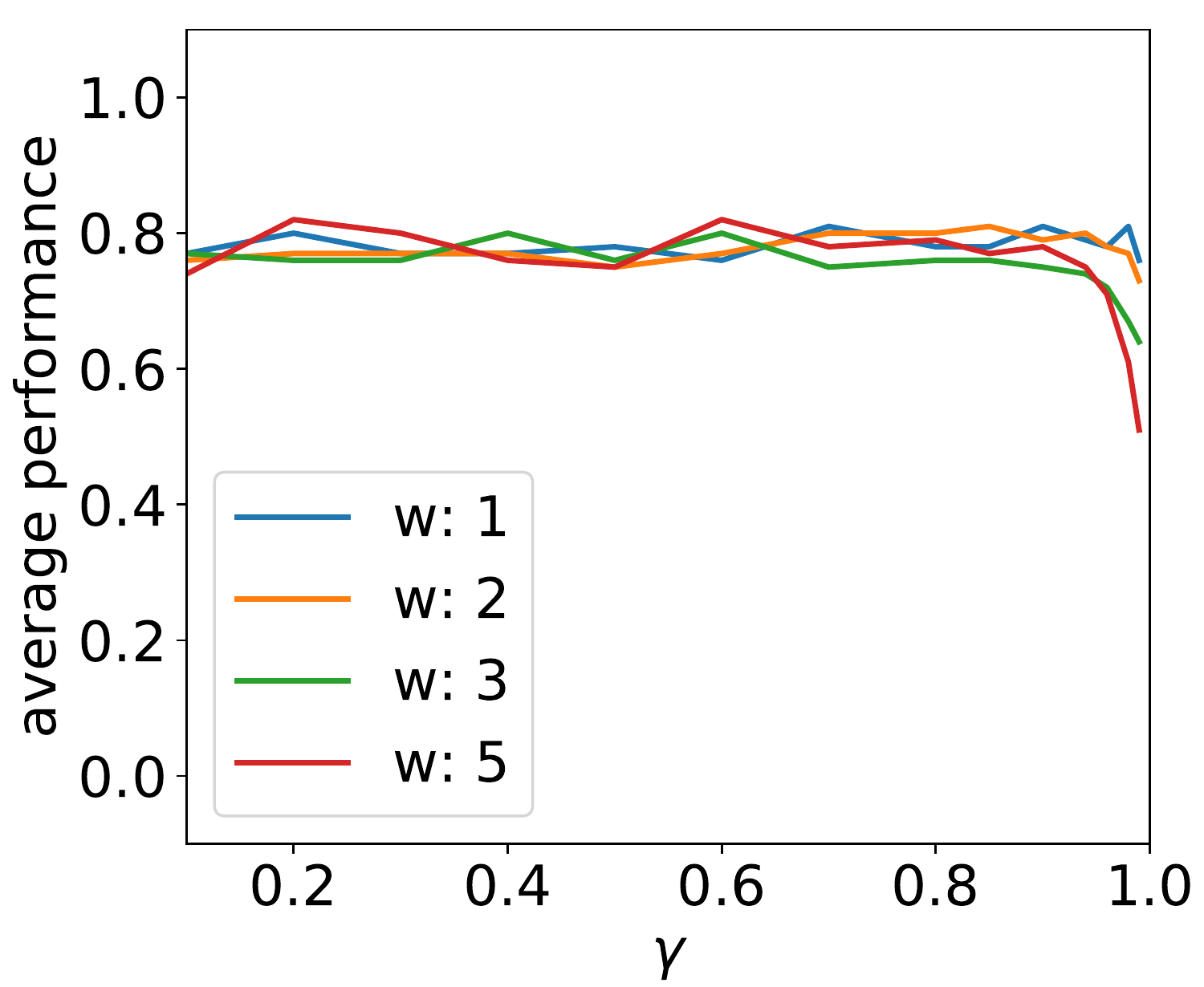}
\includegraphics[width=4.5cm]{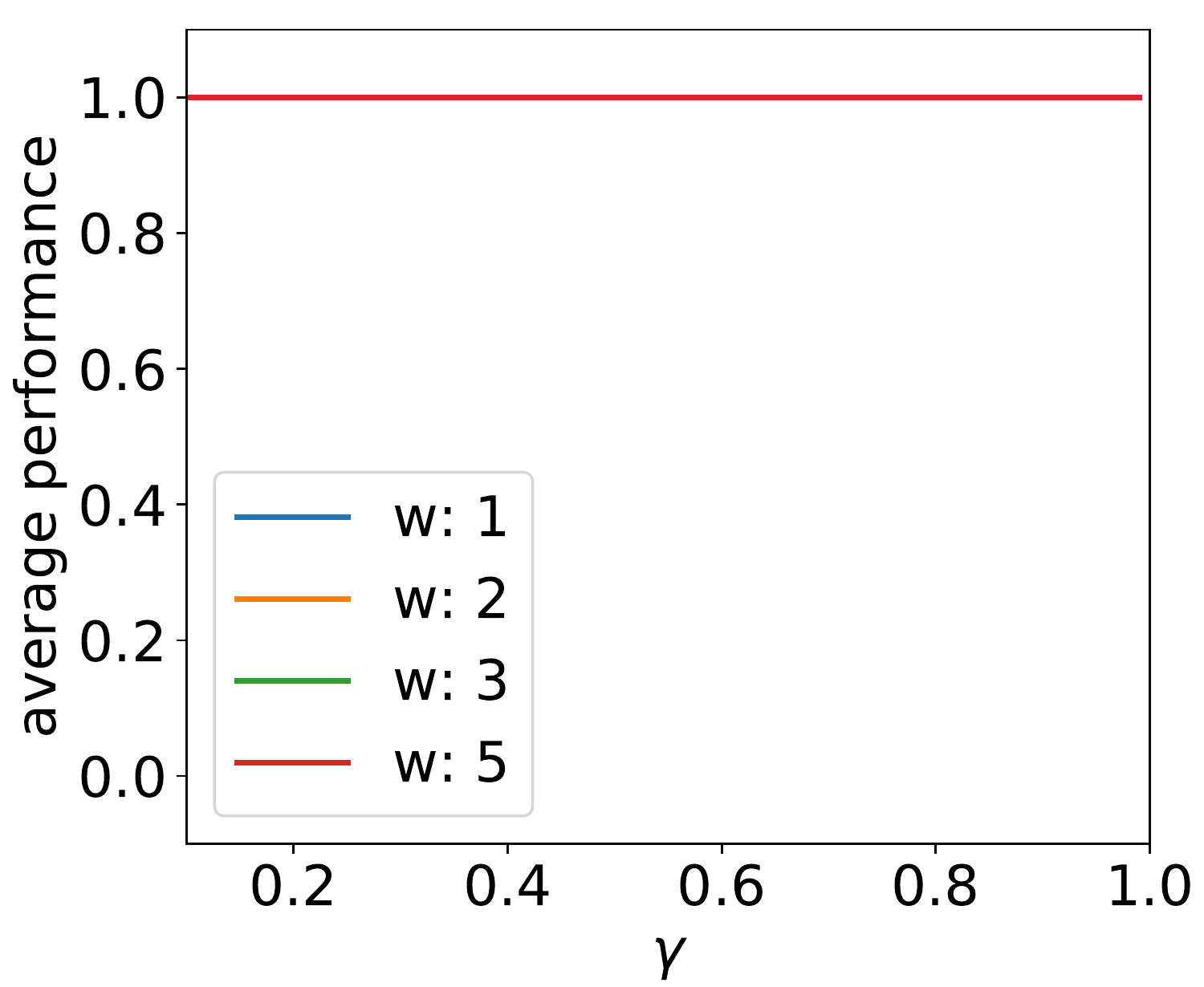}
\caption{Early performance (top) and final performance (bottom) on the chain task.}
\vspace{-15pt}
\label{fig:reg result 2}
\end{wrapfigure}

We test our method by returning to the full version of the chain task and the same performance metric $F$ as used in Section \ref{sec:optimization}, which measures whether or not the greedy policy is optimal. We used $k=200$, $\beta_{reg}=0.1$, and $\beta_{log}=0.01$  (the value of  $\beta_{reg}\cdot \beta_{log}$ is equal to the value of $\alpha$ used in Section \ref{sec:optimization}).
Figure~\ref{fig:reg result 2} plots the result for early learning as well as the final performance. Comparing these graphs with the graphs from Figure~\ref{fig:reg result} shows that logarithmic Q-learning has successfully resolved the optimization issues of regular Q-learning related to the use of low discount factors in conjunction with function approximation. 

Combined with the observation from Section \ref{sec:metric gap effect} that the best discount factor is task-dependent, and the convergence proof, which guarantees that logarithmic Q-learning converges to the same policy as regular Q-learning, these results demonstrate that logarithmic Q-learning is able to solve tasks that are challenging to solve with Q-learning. Specifically, if a finite-horizon performance metric is used and the task is such that the metric gap is substantially smaller for lower discount factors, but performance falls flat for these discount factors due to function approximation.

Finally, we test our approach in a more complex setting by comparing the performance of DQN \citep{mnih:nature15} with a variant of it that implements our method, which we will refer to as LogDQN.\footnote{The code for the experiments can be found at: \url{https://github.com/microsoft/logrl} } To enable easy baseline comparisons, we used the Dopamine framework for our experiments \citep{dopamine}. This framework not only contains open-source code of several important deep RL methods, but also contains the results obtained with these methods for a set of 60 games from the Arcade Learning Environment \citep{bellemare2013arcade, machado2018revisiting}. This means that direct comparison to some important baselines is possible.

%In order to adapt DQN's model to provide estimates of both $\tQ^+$ and $\tQ^-$, the final output layer is doubled, and half of it is used to estimate $\tQ^+$, while the other half estimates $\tQ^-$. All the other layers are shared between $\tQ^+$ and $\tQ^-$ and remain the same as baseline DQN's. Because both $\tQ^+$ and $\tQ^-$ are updated using the same samples, the replay memory does not require modification, so the memory footprint does not change. Furthermore, because $\tQ^+$ and $\tQ^-$ are updated simultaneously using a single pass through the model, the computational cost of LogDQN and DQN are similar. %Further implementation details are provided in the supplementary material. 

Our LogDQN implementation consists of a modification of the Dopamine's DQN code. Specifically, in order to adapt DQN's model to provide estimates of both $\tQ^+$ and $\tQ^-$, the final output layer is doubled in size, and half of it is used to estimate $\tQ^+$ while the other half estimates $\tQ^-$. All the other layers are shared between $\tQ^+$ and $\tQ^-$ and remain unchanged. Because both $\tQ^+$ and $\tQ^-$ are updated using the same samples, the replay memory does not require modification, so the memory footprint does not change. Furthermore, because $\tQ^+$ and $\tQ^-$ are updated simultaneously using a single pass through the model, the computational cost of LogDQN and DQN is similar. Further implementation details are provided in Appendix C.

The published Dopamine baselines are obtained on a stochastic version of Atari using \emph{sticky actions} \citep{machado2018revisiting} where with 25\% probability the environment executes the action from the previous time step instead of the agent's new action. Hence, we conducted all our LogDQN experiments on this stochastic version of Atari as well.

While Dopamine provides baselines for 60 games in total, we only consider the subset of 55 games for which human scores have been published, because only for these games a `human-normalized score' can be computed, which is defined as:
\begin{equation}
\frac{\text{Score}_\text{Agent} - \text{Score}_\text{Random}}{\text{Score}_\text{Human} - \text{Score}_\text{Random}}\,.
\label{eq:normalization}
\end{equation}
We use Table 2 from \cite{wang2016dueling} to retrieve the human and random scores.

%The published Dopamine  baselines are obtained on a stochastic version of Atari %\citep{bellemare2013arcade} using \emph{sticky actions} \citep{machado2018revisiting}, where with 25\% probability the environment executes the action from the previous time step instead of the agent's new action.
%Hence, we conducted our LogDQN experiments on this stochastic version of Atari as well. Figure \ref{fig:atari results} compares the performance of LogDQN with the published DQN baseline. 

We optimized hyper-parameters using a subset of 6 games. In particular, we performed a scan over the discount factor $\gamma$ between $\gamma=0.84$ and $\gamma=0.99$. For DQN, $\gamma=0.99$ was optimal; for LogDQN, the best value in this range was $\gamma=0.96$. We tried lower $\gamma$ values as well, such as $\gamma=0.1$ and $\gamma=0.5$, but this did not improve the overall performance over these 6 games. For the other hyper-parameters of LogDQN we used $k=100$, $c = 0.5$, $\beta_{log} = 0.0025$, and $\beta_{reg} = 0.1$. The product of $\beta_{log}$ and $\beta_{reg}$ is $0.00025$, which is the same value as the (default) step-size $\alpha$ of DQN.  We used different values of $d$ for the positive and negative heads: we set $d$ based on (\ref{eq:d setting}) with $q_{init} = 1$ for the positive head, and $q_{init} = 0$ for the negative head.
Results from the hyper-parameter optimization, as well as further implementation details are provided
in Appendix C. 

%We used $\gamma = 0.96, k = 100, c = 0.5, \beta_{log} = 0.0025$, and $\beta_{reg} = 0.1$. We used different values of $d$ for the positive and negative head: we set $d$ based on (\ref{eq:d setting}) with $q_{init} = 1$ for the positive head, and $q_{init} = 0$ for the negative head. DQN uses $\gamma = 0.99$, and $\alpha = 0.00025$ (this is equivalent to the effective step-size for LogDQN, as $\alpha = \beta_{log}\cdot\beta_{reg}$
%We tried lower $\gamma$ values as well, such as $\gamma=0.1$ and $\gamma=0.5$, but this did not improve the overall performance over these 6 games. In the final section, we discuss what the reason for this could be.). 

\begin{wrapfigure}{}{0.32\textwidth}
\vspace{-15pt}
\centering
\includegraphics[width=4.5cm]{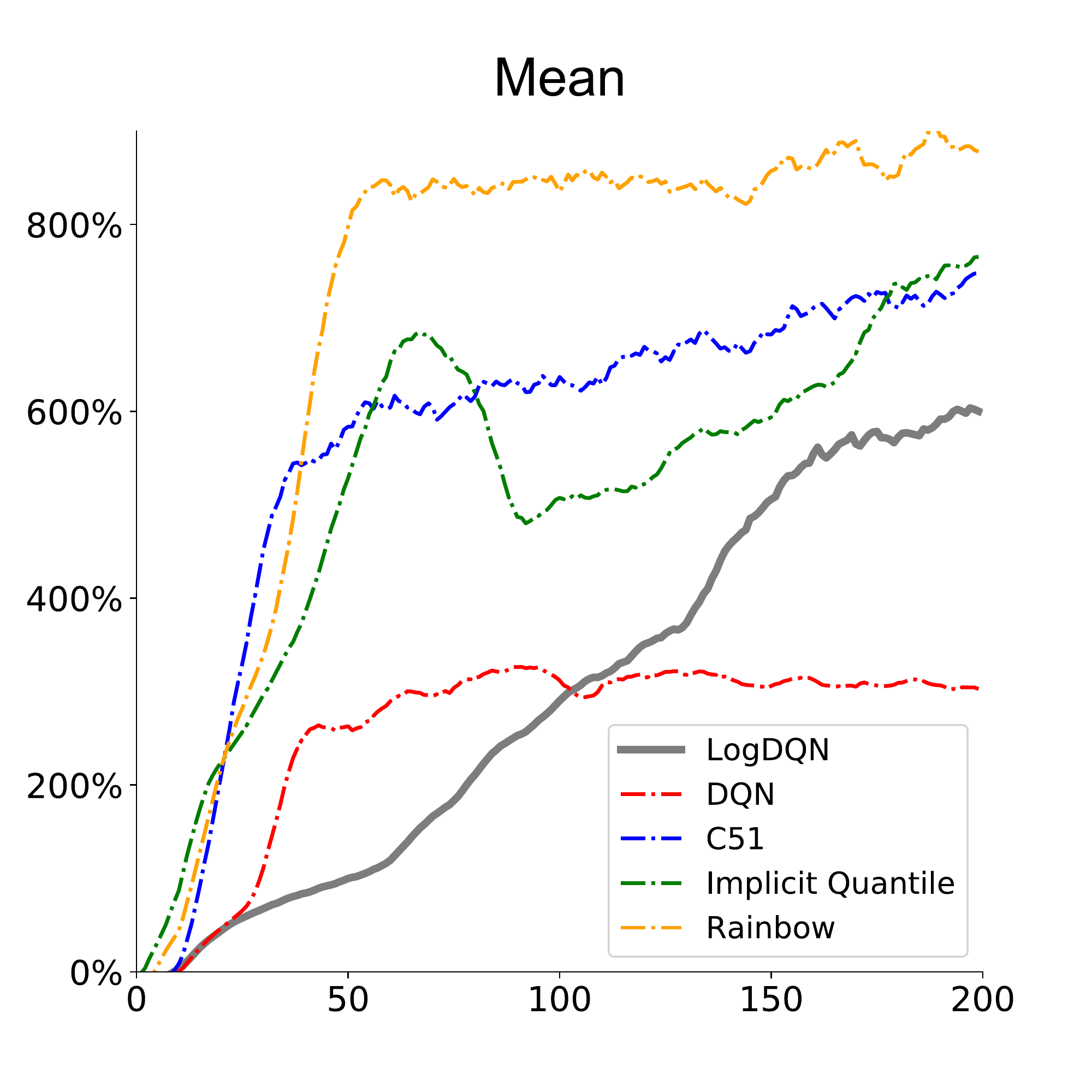}
\includegraphics[width=4.5cm]{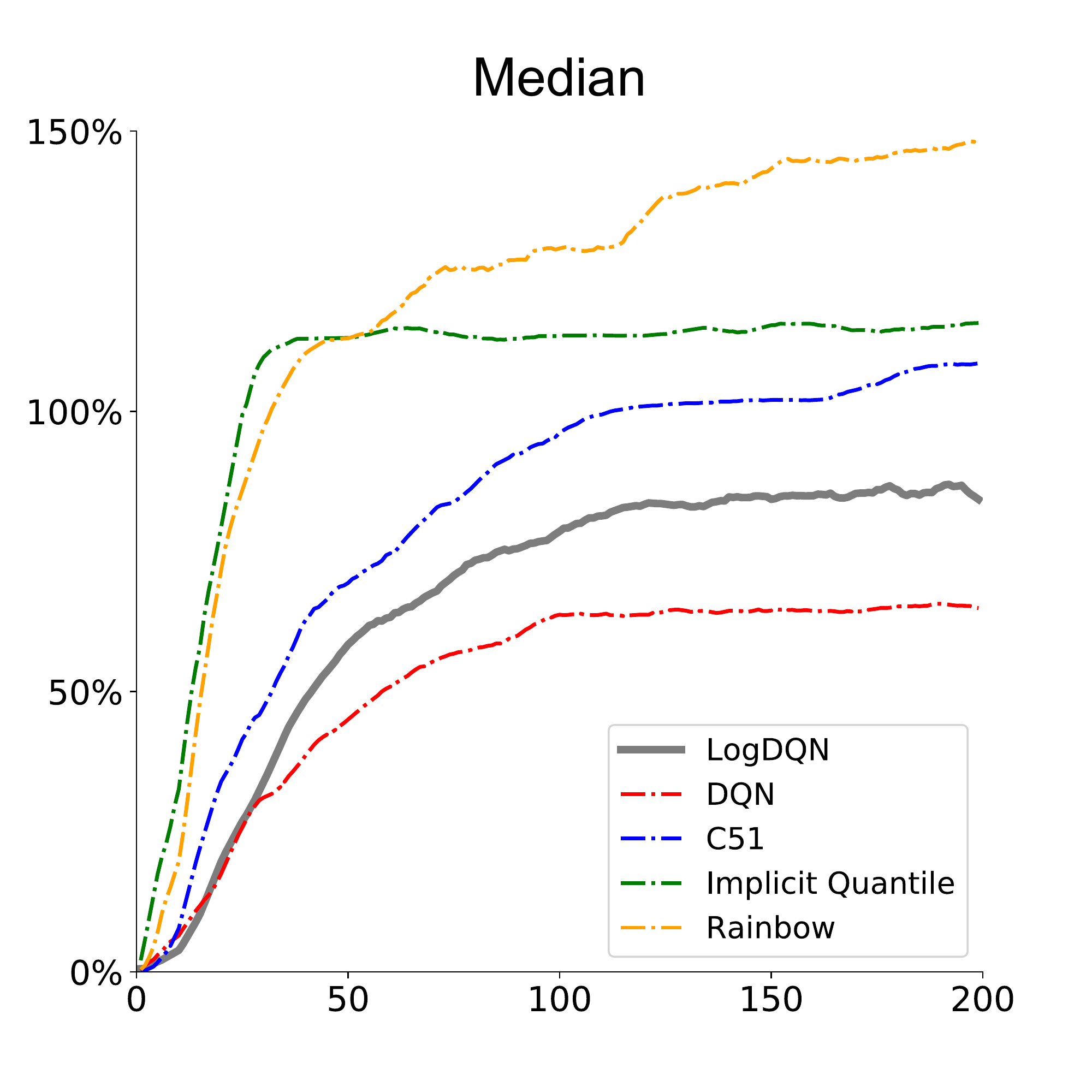}
\caption{Human-normalized mean (left) and median (right) scores on 55 Atari games for LogDQN and various other algorithms.}
%\vspace{-15pt}
\label{fig:mean_median}
\end{wrapfigure}

% \begin{figure}[tbh]
% \begin{center}
% \includegraphics[width=0.4\columnwidth]{img/mean2.pdf}
% \includegraphics[width=0.4\columnwidth]{img/median2.pdf}
% \caption{Human-normalized mean (left) and median (right) scores on 55 Atari games for LogDQN and various other algorithms.}
% \label{fig:mean_median}
% \end{center}
% \end{figure}

\begin{figure}[t]
\begin{center}
\resizebox{13.8cm}{4.2cm}{
\includegraphics[width=13.8cm]{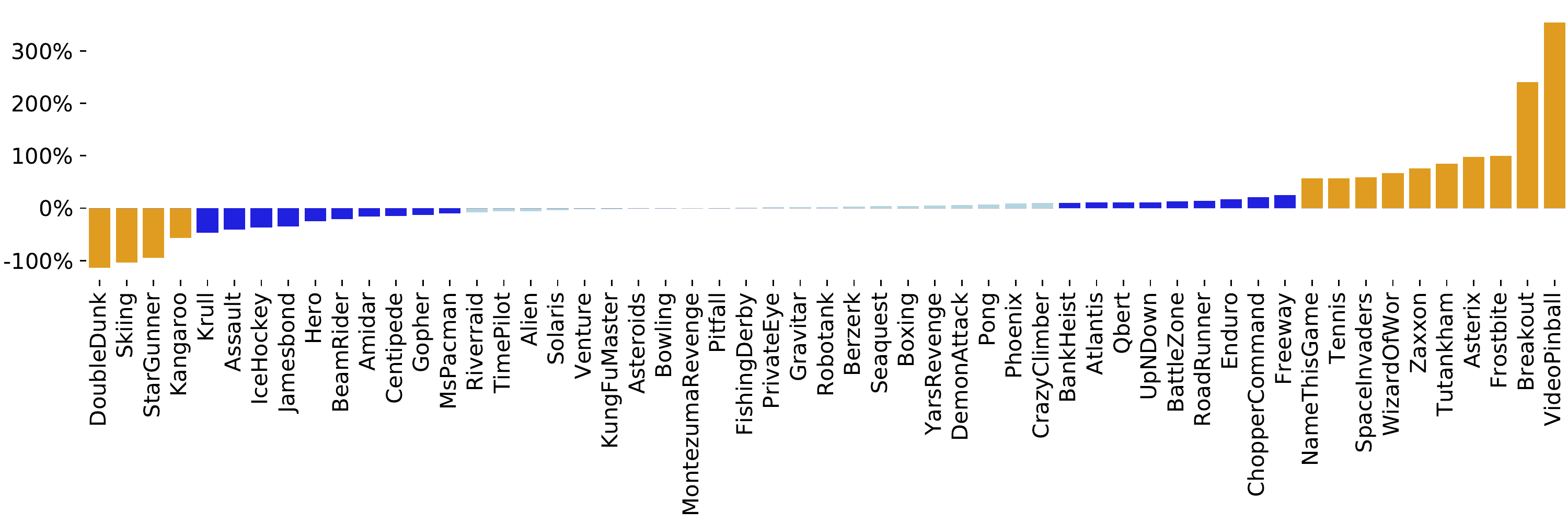}}
\caption{Relative performance of LogDQN w.r.t. DQN (positive percentage means LogDQN outperforms DQN). Orange bars indicate a performance difference larger than 50\%; dark-blue bars indicate a performance difference between 10\% and 50\%; light-blue bars indicate a performance difference smaller than 10\%. }
\label{fig:atari results} 
\end{center}
\vspace{-0.2cm}
\end{figure}

Figure~\ref{fig:atari results} shows the performance of LogDQN compared to DQN per game, using the same comparison equation as used by \citet{wang2016dueling}:
\begin{displaymath}
\frac{\text{Score}_\text{LogDQN} - \text{Score}_\text{DQN}}{\max(\text{Score}_\text{DQN}
,\text{Score}_\text{Human}) - \text{Score}_\text{Random}}\,.
\end{displaymath}
where $\text{Score}_\text{LogDQN/DQN}$ is computed by averaging over the last 10\% of each learning curve (i.e., last 20 epochs).

Figure~\ref{fig:mean_median} shows the mean and median of the human-normalized score of LogDQN, as well as DQN. We also plot the performance of the other baselines that Dopamine provides: C51 \citep{bellemare2017:c51}, Implicit Quantile Networks \citep{dabney2018:iqn}, and Rainbow \citep{hessel2018:rainbow}. These baselines are just for reference; we have not attempted to combine our technique with the techniques that these other baselines make use of.

\vspace{-0.1cm}
\vspace{-\spaceSec em}
\section{Discussion and Future Work}
\vspace{-\spaceSec em}
\vspace{-0.1cm}

% basis for hypothesis
Our results provide strong evidence for our hypothesis that large differences in action-gap sizes are detrimental to the performance of approximate RL. A possible explanation could be that optimizing on the $L_2$-norm (\ref{eq:regular loss function}) might drive towards an average squared-error that is similar across the state-space. However, the error-landscape required to bring the approximation error below the action-gap across the state-space has a very different shape if the action-gap is orders of magnitude different in size across the state-space. This mismatch between the required error-landscape and that produced by the $L_2$-norm might lead to an ineffective use of the function approximator. Further experiments are needed to confirm this.

The strong performance we observed for $\gamma=0.96$ in the deep RL setting is unlikely solely due to a difference in metric gap. We suspect that there are also other effects at play that make LogDQN as effective as it is. On the other hand, at (much) lower discount factors, the performance was not as good as it was for the high discount factors. 
%We had expected that at least for some of the 6 games, a small discount factor (i.e., somewhere between $\gamma=0.1$ and $\gamma=0.5$) would be better.  
We believe a possible reason could be that since such low values are very different than the original DQN settings, some of the other DQN hyper-parameters might no longer be ideal in the low discount factor region. An interesting future direction would be to re-evaluate some of the other hyper-parameters in the low discount factor region.  

\subsubsection*{Acknowledgments}

We like to thank Kimia Nadjahi for her contributions to a convergence proof of an early version of logarithmic Q-learning. This early version ultimately was replaced by a significantly improved version that required a different convergence proof.

%\bibliography{references}
%\bibliographystyle{unsrtnat}

\newpage
\appendix

\section{Proof of Convergence for Logarithmic Q-learning}
\label{sec:convergence}

\subsection{Definitions and Theorem}
\label{sec:definitions}

Our logarithmic Q-learning method is defined by the following equations:

\begin{equation}
f(x) := c\,\ln(x + \gamma^k) + d\quad ; \quad f^{-1}(x) := e^{(x - d)/c} - \gamma^k
\label{eq:f}
\end{equation}

\begin{equation}
r^+_t:=
\begin{cases}
r_t & \mbox{if } r_t \geq 0\\
0  & \mbox{otherwise }\ \\
\end{cases}
\quad ; \quad
r^-_t :=
\begin{cases}
|r_t| & \mbox{if } r_t < 0\\
0  & \mbox{otherwise }\ \\
\end{cases}
\label{eq:reward decomposition2}
\end{equation}

\begin{equation}
Q_t(s,a) := f^{-1}\left(\tQ_{t}^+(s,a)\right) - f^{-1}\left(\tQ_{t}^-(s,a)\right)
\label{eq:Q definition2}
\end{equation}

\begin{equation}
\tilde a_{t+1} := \arg\max_{a'} \Big( Q_t(s_{t+1},a')\Big)
\label{eq: a_tilde}
\end{equation}

\begin{equation}
U^+_t :=   r^+_{t} + \gamma f^{-1}\left( \tQ^+_t(s_{t+1},\tilde a_{t+1})\right)
\label{eq:update_target_plus}
\end{equation}

\begin{equation}
\hat U^+_t :=   f^{-1}\left(\tQ^+_t(s_t, a_t)\right) +
\beta_{reg,t} \left(U^+_t -  f^{-1}\left(\tQ^+_t(s_t, a_t)\right) \right)
\label{eq:reg_update_plus}
\end{equation}

\begin{equation}
\tQ_{t+1}^+(s_t, a_t) :=  \tQ_t^+(s_t, a_t) +
\beta_{log,t} \left(f\left( \hat U_t^+ \right) - \tQ^+_t(s_t, a_t)\right)
\label{eq:log_update_plus}
\end{equation}

\begin{equation}
U^-_t :=   r^-_{t} + \gamma f^{-1}\left( \tQ^-_t(s_{t+1},\tilde a_{t+1})\right)
\label{eq:update_target_min}
\end{equation}

\begin{equation}
\hat U^-_t :=   f^{-1}\left(\tQ^-_t(s_t, a_t)\right) +
\beta_{reg,t} \left(U^-_t -  f^{-1}\left(\tQ^-_t(s_t, a_t)\right) \right)
\label{eq:reg_update_min}
\end{equation}

\begin{equation}
\tQ_{t+1}^-(s_t, a_t) :=  \tQ_t^-(s_t, a_t) +
\beta_{log,t} \left(f\left( \hat U_t^- \right) - \tQ^-_t(s_t, a_t)\right)
\label{eq:log_update_min}
\end{equation}

For these equations, the following theorem holds:
\begin{theorem}
Under the definitions above, $Q_t$ converges to $Q^*$ w.p. 1 if the following conditions hold: 
\begin{enumerate}
    \item $0\leq \beta_{log,t} \cdot \beta_{reg,t} \leq 1$
    \item $\sum_{t=0}^\infty  \beta_{log,t}\cdot \beta_{reg,t} = \infty$
    \item $\sum_{t=0}^\infty (\beta_{log,t}\cdot \beta_{reg,t})^2 < \infty$
    \item $\lim_{t\rightarrow\infty} \beta_{reg,t} = 0$
\end{enumerate}
\end{theorem}

\subsection{Proof - part 1}

We define $Q^+_t(s,a) := f^{-1}(\tQ^+_t(s,a))$ and prove in part 2 that from (\ref{eq:log_update_plus}), (\ref{eq:reg_update_plus}), and (\ref{eq:update_target_plus}) it follows that:
\begin{equation}
Q_{t+1}^+(s_t, a_t) =  Q_t^+(s_t, a_t) + \beta_{reg,t}\cdot\beta_{log,t}  \left(U_t^+ - Q^+_t(s_t, a_t)  + c_t^+\right)\,,
\label{eq:Q_plus}
\end{equation}
with $c_t^+$ converging to zero w.p. 1 under condition 4 of the theorem, and $U^+_t$ defined as:
$$U^+_t :=   r^+_{t} + \gamma \, Q^+_t(s_{t+1},\tilde a_{t+1})\,.$$
Similarly, using definition $Q^-_t(s,a) := f^{-1}(\tQ^-_t(s,a))$ and (\ref{eq:log_update_min}), (\ref{eq:reg_update_min}), and (\ref{eq:update_target_min}) it follows that:
\begin{equation}
Q_{t+1}^-(s_t, a_t) =  Q_t^-(s_t, a_t) + \beta_{reg,t}\cdot\beta_{log,t}  \left(U_t^- - Q^-_t(s_t, a_t)  + c_t^-\right)\,,
\label{eq:Q_min}
\end{equation}
with $c_t^-$ converging to zero w.p. 1 under condition 4 of the theorem, and $U_t^-$ defined as:
$$U^-_t :=   r^+_{t} + \gamma \, Q^-_t(s_{t+1},\tilde a_{t+1})\,.$$
It follows directly from the definitions of $Q^+_t$ and $Q^-_t$ and (\ref{eq:Q definition2}) that:
\begin{equation}
Q_t(s,a) = Q_t^+(s,a) - Q_t^{-}(s,a)\,.
\label{eq:Q_min_plus}
\end{equation}
Subtracting  (\ref{eq:Q_min}) from (\ref{eq:Q_plus}) and substituting this equivalence yields:
$$Q_{t+1}(s_t, a_t) =  Q_t(s_t, a_t) + \beta_{reg,t}\cdot\beta_{log,t}  \left(r_t^+ - r_t^- + \gamma Q_t(s_{t+1},\tilde a_{t+1}) - Q_t(s_t, a_t)  + c_t^+  - c_t^-\right)\,.$$
From (\ref{eq:reward decomposition2}) it follows that $r_t = r_t^+ - r_t^-$. Furthermore, the following holds:
\begin{eqnarray*}
Q_t(s_{t+1},\tilde a_{t+1}) &=& Q_t\left(s_{t+1},\arg\max_{a'} \big( Q_t(s_{t+1},a')\big) \right)\\
&=& \max_{a'} Q_t\big(s_{t+1},a'\big)
\end{eqnarray*}
Using these equivalences and defining $\alpha_t := \beta_{reg,t}\cdot\beta_{log,t}$ and $c_t :=c_t^+ - c_t^-$, it follows that:
\begin{equation}
Q_{t+1}(s_t, a_t) =  Q_t(s_t, a_t) + \alpha_t \left(r_t + \gamma \max_{a'} Q_t(s_{t+1},a') - Q_t(s_t, a_t)  + c_t \right)\,,
\label{eq:Qfinal}
\end{equation}
with $c_t$ converging to zero w.p. 1 under condition 4 of the theorem. This is a noisy Q-learning algorithm with the noise term decaying to zero. As we show in part 2, $c_t$ is fully specified (in the positive case, and likewise in the negative case) by $Q_t^{+}$, $U_t^{+}$, and $\beta_{reg, t}$, which implies that $c_t$ is measurable given information at time $t$, as required by Lemma 1 in \cite{singh2000convergence}. Invoking that Lemma, it can therefore be shown that the iterative process defined by (\ref{eq:Qfinal}) converges to $Q^*_t$ if $0\leq \alpha \leq 1$, $\sum_{t=0}^\infty  \alpha = \infty$, and $\sum_{t=0}^\infty \alpha_t^2 < \infty$, as is guaranteed by the first three conditions of the theorem. The steps are similar to the proof of Theorem 1 of the same reference, which we do not repeat here.

\subsection{Proof - part 2}

In this section, we prove that (\ref{eq:Q_plus}) holds under the definitions from Section \ref{sec:definitions},  $Q^+_t(s,a) := f^{-1}(\tQ^+_t(s,a))$, and condition 4 of the theorem. The proof of (\ref{eq:Q_min}) follows the same steps, but with the `-' variants of the different variables instead. For readability, we use $\beta_1$ for $\beta_{log,t}$ and $\beta_2$ for $\beta_{reg,t}$.

The definition of $Q^+_t$ implies $\tQ^+_t(s,a) = f\left(Q^+_t(s,a)\right)$. Using these equivalences, we can rewrite (\ref{eq:log_update_plus}), (\ref{eq:reg_update_plus}), and (\ref{eq:update_target_plus}) in terms of $Q_t$:
\begin{equation}
f(Q^+_{t+1}(s_t, a_t)) =  f(Q^+_t(s_t, a_t)) +
\beta_1 \left(f( \hat U^+_t ) - f(Q^+_t(s_t, a_t))\right)\,,
\label{eq:log_update2}
\end{equation}
with
\begin{equation}
\hat U^+_t =   Q^+_t(s_t, a_t) +
\beta_2 \left(r^+_{t} + \gamma\, Q^+_t(s_{t+1}, \tilde a_{t+1}) -  Q^+_t(s_t, a_t) \right) \,.
\label{eq:ref_update2}
\end{equation}
By applying $f^{-1}$ to both sides of (\ref{eq:log_update2}), we get:
\begin{equation}
Q^+_{t+1}(s_t, a_t) =  f^{-1}\left(f(Q^+_t(s_t, a_t)) +
\beta_1 \left(f\left( \hat U^+_t \right) - f(Q^+_t(s_t, a_t))\right)\right)\,,
\label{eq:log_update3}
\end{equation}
which can be rewritten as:
\begin{equation}
Q^+_{t+1}(s_t, a_t) = Q^+_{t}(s_t, a_t) + \beta_1 \left( \hat U^+_t - Q^+_t(s_t,a_t)\right) + e^+_t\,,
\label{eq:log_update4}
\end{equation}
\newpage
where $e^+_t$ is the error due to averaging in the log-space instead of in the regular space:
\begin{multline}
e^+_t := f^{-1}\left(f(Q^+_t(s_t, a_t)) +
\beta_1 \left(f( \hat U^+_t) - f(Q^+_t(s_t, a_t))\right)\right) \\ - Q^+_{t}(s_t, a_t) - \beta_1 \left( \hat U^+_t - Q^+_t(s_t,a_t)\right)
\end{multline}

\begin{figure}[tbh]
\begin{center}
\includegraphics[height=5cm]{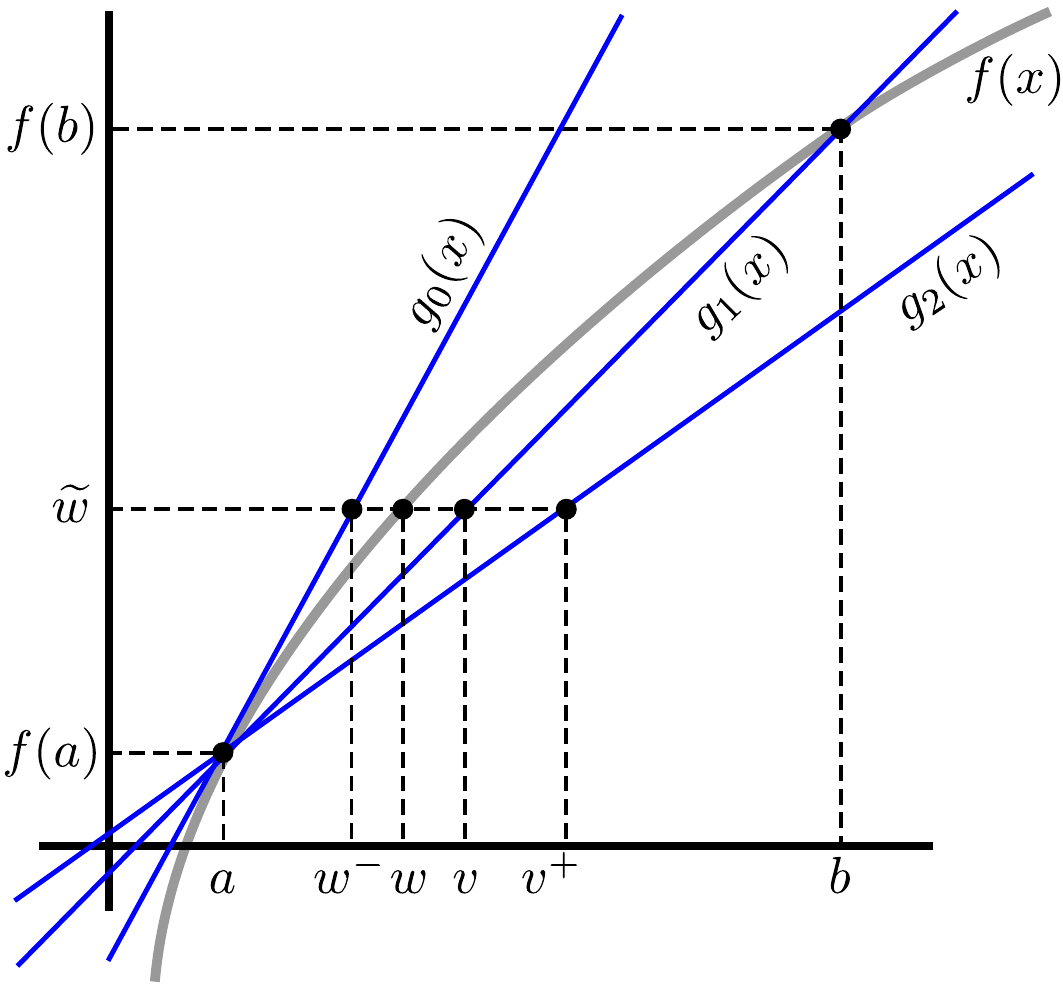}
\hspace{1cm}
\includegraphics[height=5cm]{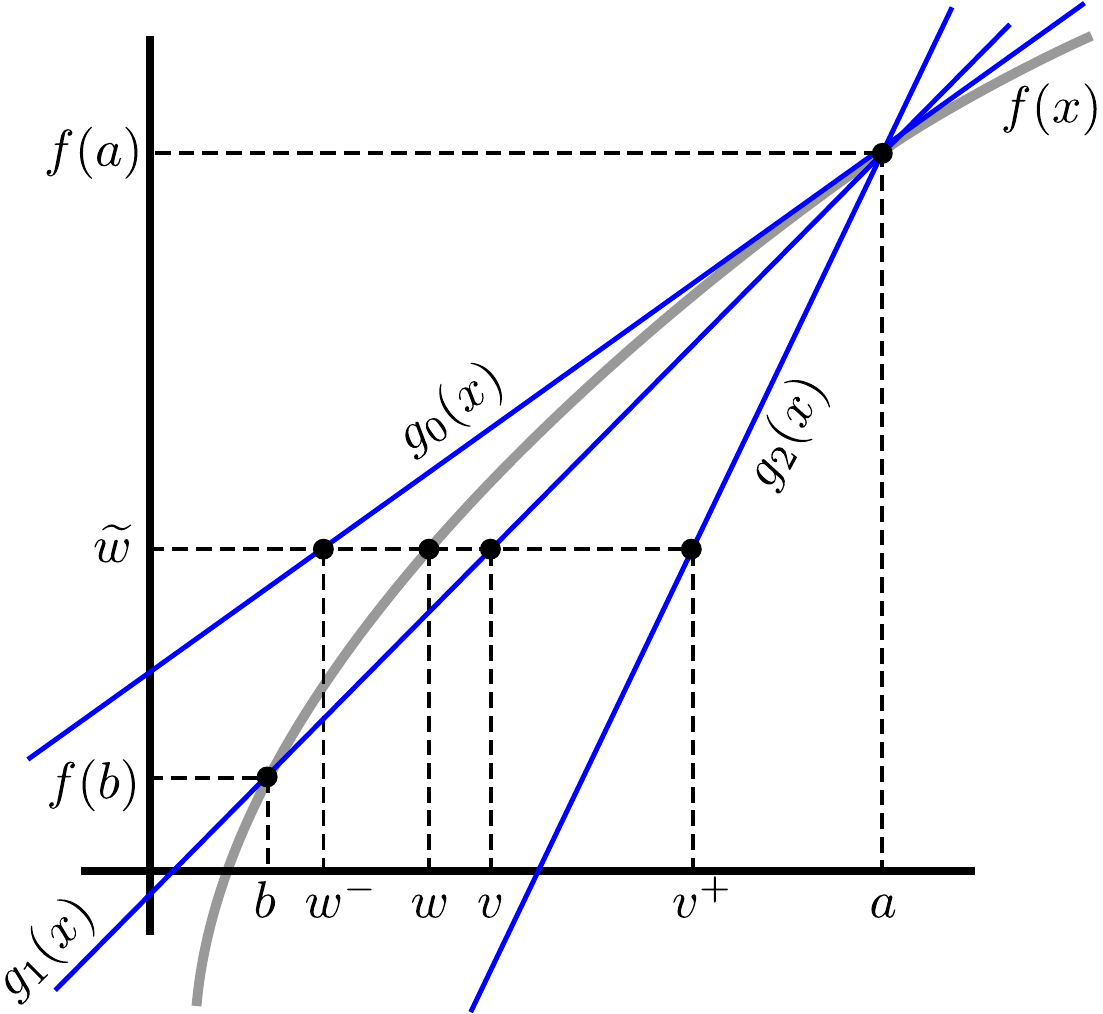}
\caption{Bounding the error, for the case $a < b$ (left) and for $a > b$ (right).}
\label{fig:error bound}
\end{center}
\end{figure}

The key to proving (\ref{eq:Q_plus}), and by extension the theorem, is proving that $e_t^+$ goes sufficiently fast to 0. We prove this by defining a bound on $|e^+_t |$ and showing that this bound goes to 0. Figure~\ref{fig:error bound} illustrates the bound. The variables in the figure refer to the following quantities:
\begin{eqnarray*}
a &\rightarrow& Q_t^+(s_t,a_t)\\
b &\rightarrow & \hat U_t^+ \\
v &\rightarrow &  (1-\beta_1)\,a + \beta_1\,b\\
\tilde w &\rightarrow & (1-\beta_1)f(a) + \beta_1 f(b)\\
w & \rightarrow & f^{-1}(\tilde w)
\end{eqnarray*}
The error $e_t^+$ corresponds with: $$e_t^+ = f^{-1}\big((1-\beta_1)f(a) + \beta_1 f(b)\big) - \big((1-\beta_1)a + \beta_1 b\,\big) = f^{-1}(\tilde w) - v = w - v$$
Note here that since $f$ is a strictly concave function, the definition of $\tilde w$ and $v$ directly imply $\tilde w < f(v)$. Because $f^{-1}$ is monotonically increasing, it follow that $w<v$, which yields $|e^{+}_{t}|=v - w$.

In both graphs of Figure~\ref{fig:error bound}, besides the mapping function $f(x)$, three more functions are plotted: $g_0(x)$, $g_1(x)$, and $g_2(x)$. These three functions are all linear functions passing through the point $(a,f(a))$. The function  $g_0(x)$ has derivative $f'(a)$, while $g_2(x)$ has derivative $f'(b)$. The function $g_1(x)$ passes through point $(b,f(b))$ as well, giving it derivative $(f(a)-f(b))/(a-b)$. 

As illustrated by the figure, $g_1(v) = \tilde w$ and $g_1^{-1}(\tilde w) = v$. Furthermore, for $x$ between $a$ and $b$ the following holds (in both cases):
$$g_0(x) \geq f(x) \geq g_1(x) \geq g_2(x)$$
And, equivalently:
$$g_0^{-1}(x) \leq f^{-1}(x) \leq g^{-1}_1(x) \leq g^{-1}_2(x)\,.$$

We bound $|e_t^+| = v - w$, by using a lowerbound $w^-$ for $w$ and an upperbound $v^+$ for $v$. Specifically, we define $w^- := g_0^{-1}(\tilde w)$ and $v^+ := g_2^{-1}(\tilde w)$, and can now bound the error as follows: $|e_t^+| <=  v^+ - w^-$. Next, we compute an expression for the bound in terms of $a$, $b$, and $f$.

First, note that for the derivatives of $g_0$ and $g_2$ the following holds:
$$g_0'(x) = f'(a) = \frac{f(a) - \tilde w}{a-w^-}\quad;\quad g_2'(x)= f'(b) = \frac{f(a) - \tilde w}{a-v^+}\,.$$
From this it follows that:
$$w^- = \frac{\tilde w - f(a)}{f'(a)} + a\quad;\quad v^+= \frac{\tilde w - f(a)} {f'(b)} + a\,.$$
Using this, we rewrite our bound as:
\begin{eqnarray*}
v^+ - w^- &=& \frac{\tilde w - f(a)} {f'(b)} - \frac{\tilde w - f(a)}{f'(a)} \\
&=& \left( \frac{1}{f'(b)} - \frac{1}{f'(a)} \right)\cdot (\tilde w - f(a)) \\
&=& \left( \frac{1}{f'(b)} - \frac{1}{f'(a)} \right)\cdot \Big((1-\beta_1)f(a) + \beta_1 f(b) - f(a)\Big)\\
&=& \beta_1 \left( \frac{1}{f'(b)} - \frac{1}{f'(a)} \right) \Big(f(b) - f(a) \Big)
\end{eqnarray*}
Recall that $f(x) := c\,\ln(x + \gamma^k) + d$. The derivative of $f(x)$ is 
$$f'(x) = \frac{c}{x + \gamma^k}$$
Substituting $f(x)$ and $f'(x)$ in the expression for the bound gives:
\begin{eqnarray*}
v^+ - w^- &=& \beta_1 \left( \frac{b+\gamma^k}{c} - \frac{a + \gamma^k}{c}\right)\big( c\ln(b + \gamma^k) + d - (c\ln(a + \gamma^k) + d)\big)\\
&=& \beta_1(b-a)(\ln(b + \gamma^k) - \ln(a+\gamma^k))\\
&=& \beta_1(a-b)(\ln(a + \gamma^k) - \ln(b+\gamma^k))\\
&=& \beta_1(a-b)\ln\left(\frac{a + \gamma^k}{b+\gamma^k}\right)\\
&=& \beta_1(a-b)\ln\left(\frac{a - b}{b+\gamma^k} + 1\right)
% &\leq & \beta_1(a-b)\ln\left(\frac{a - b}{\gamma^k} + 1\right)
\end{eqnarray*}
Using the definitions of $a$ and $b$, the results for the bound for $e_t^+$:
\begin{equation}
|e_t^+| \leq v^+ - w^- \leq \beta_1(Q_t^+(s_t,a_t) - \hat U_t^+ )\ln\left(\frac{Q_t^+(s_t,a_t)-\hat U_t^+}{\hat U^+_t + \gamma^k} + 1\right)
\label{eq:error bound}
\end{equation}

Definition (\ref{eq:reg_update_plus}) can be written as:
\begin{equation}
\hat U^+_t :=   Q^+_t(s_t, a_t) +
\beta_{reg,t} \left(U^+_t -  Q^+_t(s_t, a_t) \right)
\label{eq:reg_update_plus2}
\end{equation}
yielding:
\begin{eqnarray*}
Q_t^+(s_t,a_t) - \hat U^+_t &=& Q_t^+(s_t,a_t) - \Big( Q^+_t(s_t, a_t) +
\beta_{reg,t} \left(U^+_t -  Q^+_t(s_t, a_t) \right) \Big)\\
&=& \beta_2 \left( Q^{+}_t(s_t, a_t) - U^{+}_t\right)
\end{eqnarray*}
Substituting this in (\ref{eq:error bound}) gives:
$$ |e_t^+| \leq \beta_1\beta_2\big(Q^+_t(s_t, a_t) - U^+_t \big)\ln\left(\frac{\beta_2 \left( Q^+_t(s_t, a_t) - U^+_t\right)}{\hat U^+_t +\gamma^k} + 1\right)$$
Let us define $c_t^+$ as:
$$c_t^+ :=  \big(Q^+_t(s_t, a_t) - U^+_t \big)\ln\left(\frac{\beta_2 \left( Q^+_t(s_t, a_t) - U^+_t\right)}{\hat U^+_t +\gamma^k} + 1\right)$$
Hence, $|e^{+}_{t}|\le \beta_{1}\beta_{2}c^{+}_{t}$. Substituting maximum bound of $|e^{+}_{t}|$ and (\ref{eq:reg_update_plus2}) in (\ref{eq:log_update4}), we get:
\begin{equation}
Q^+_{t+1}(s_t, a_t) = Q^+_{t}(s_t, a_t) + \beta_1\beta_2 \big( U^+_t - Q^+_t(s_t,a_t)  + c_t^+ \big)
\end{equation}
with $c_t^+$ going to 0, if $\beta_2$ goes to 0, which concludes part 2 of the proof.

\section{Hypothesis Testing}

The following hypotheses are tested: 1) lower discount factors cause poor performance because they result in smaller action gaps; 2) lower discount factors cause poor performance because they result in smaller relative action gaps (i.e, the action gap of a state divided by the maximum action-value of that state). 

\begin{figure}[tbh]
\begin{center}
\includegraphics[height=3.2cm]{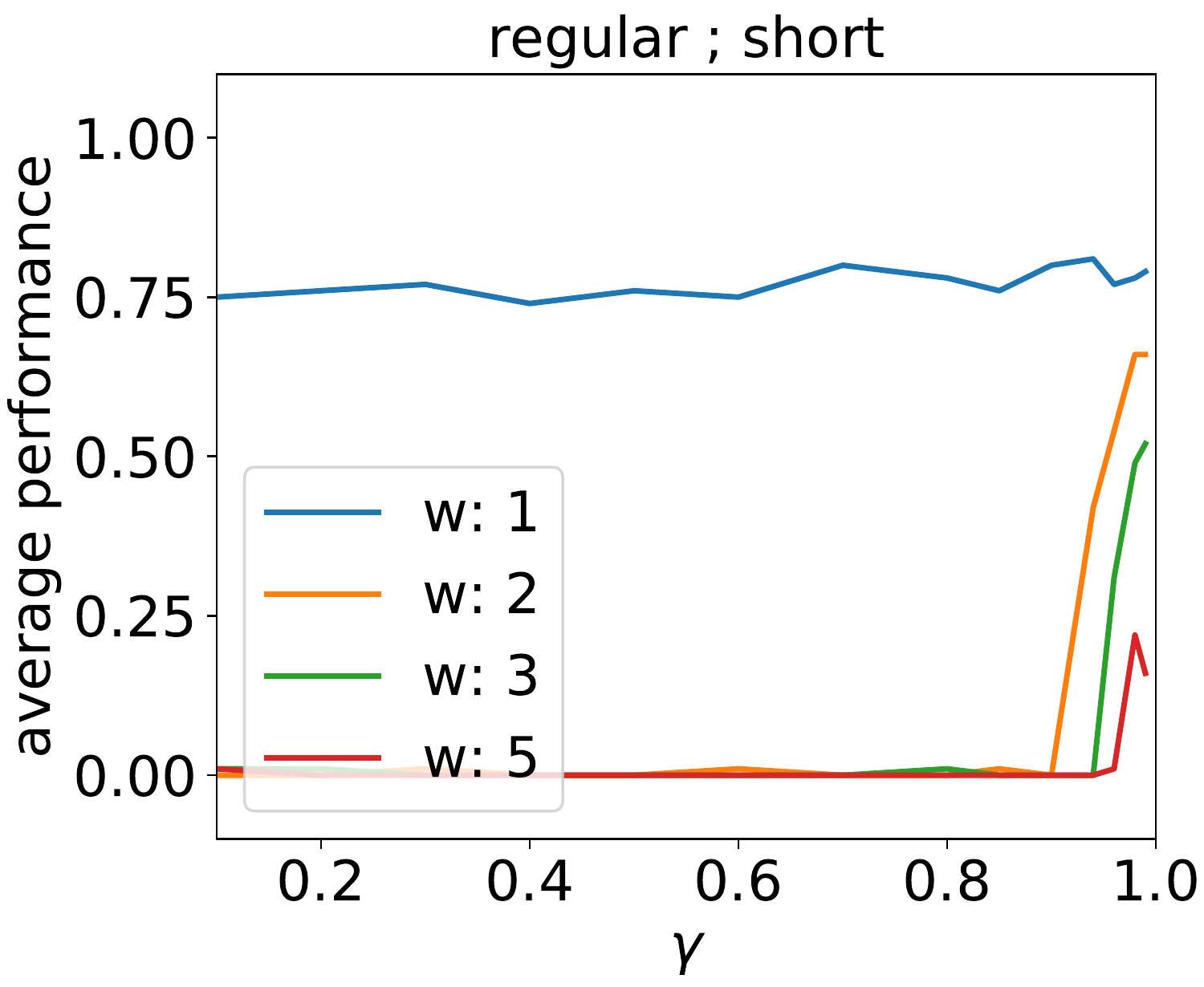}
\includegraphics[height=3.2cm]{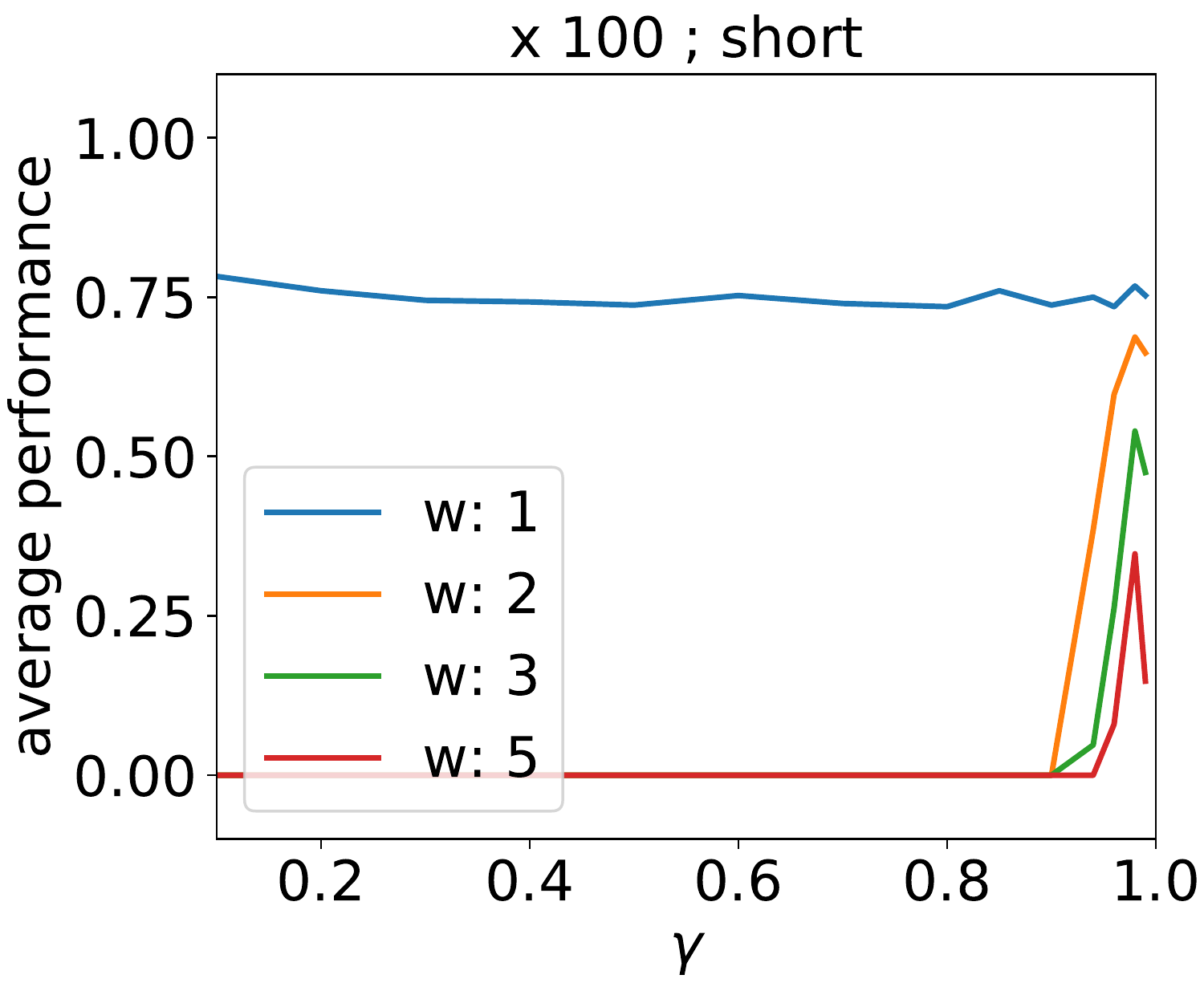}
\includegraphics[height=3.2cm]{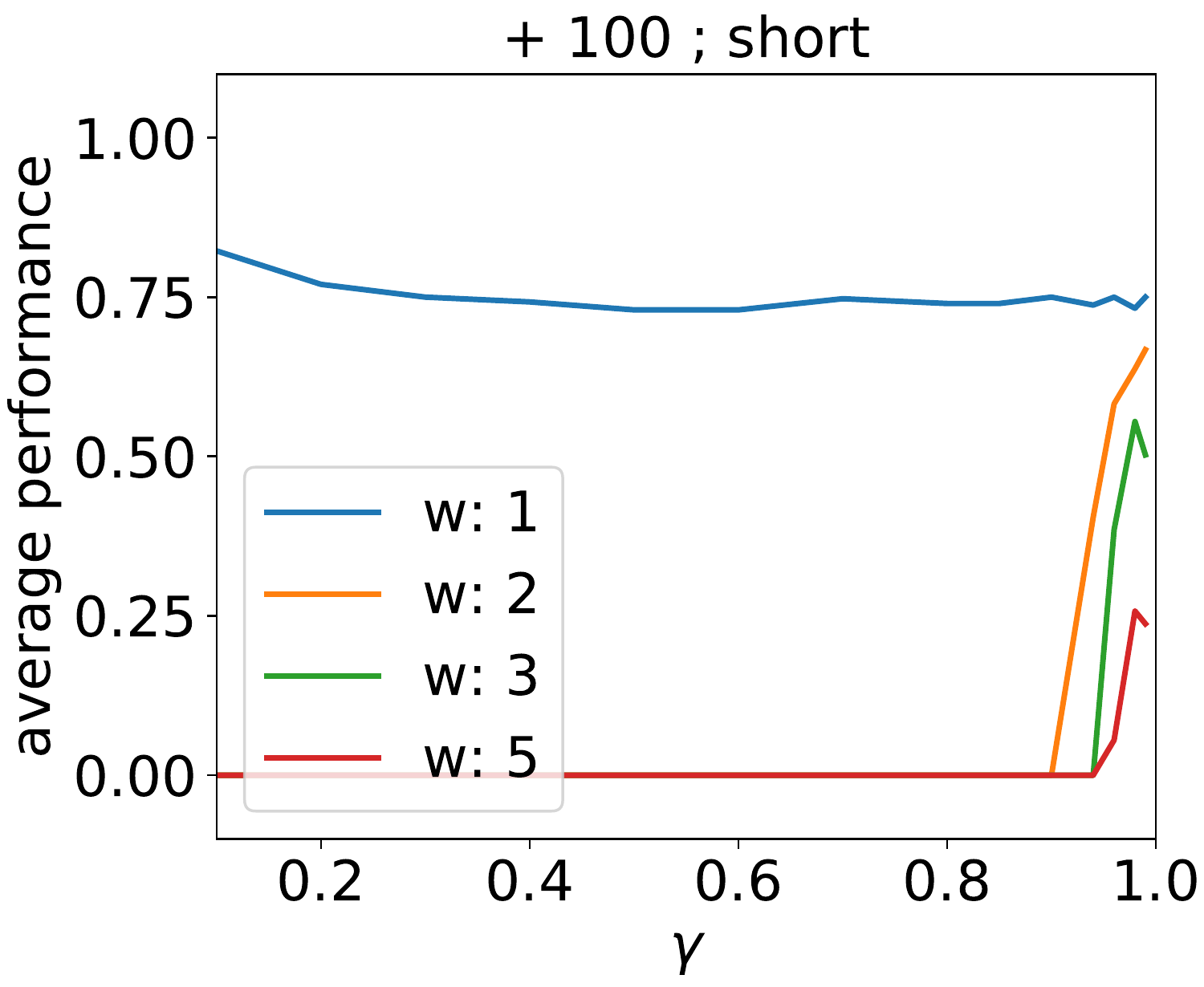}
\includegraphics[height=3.2cm]{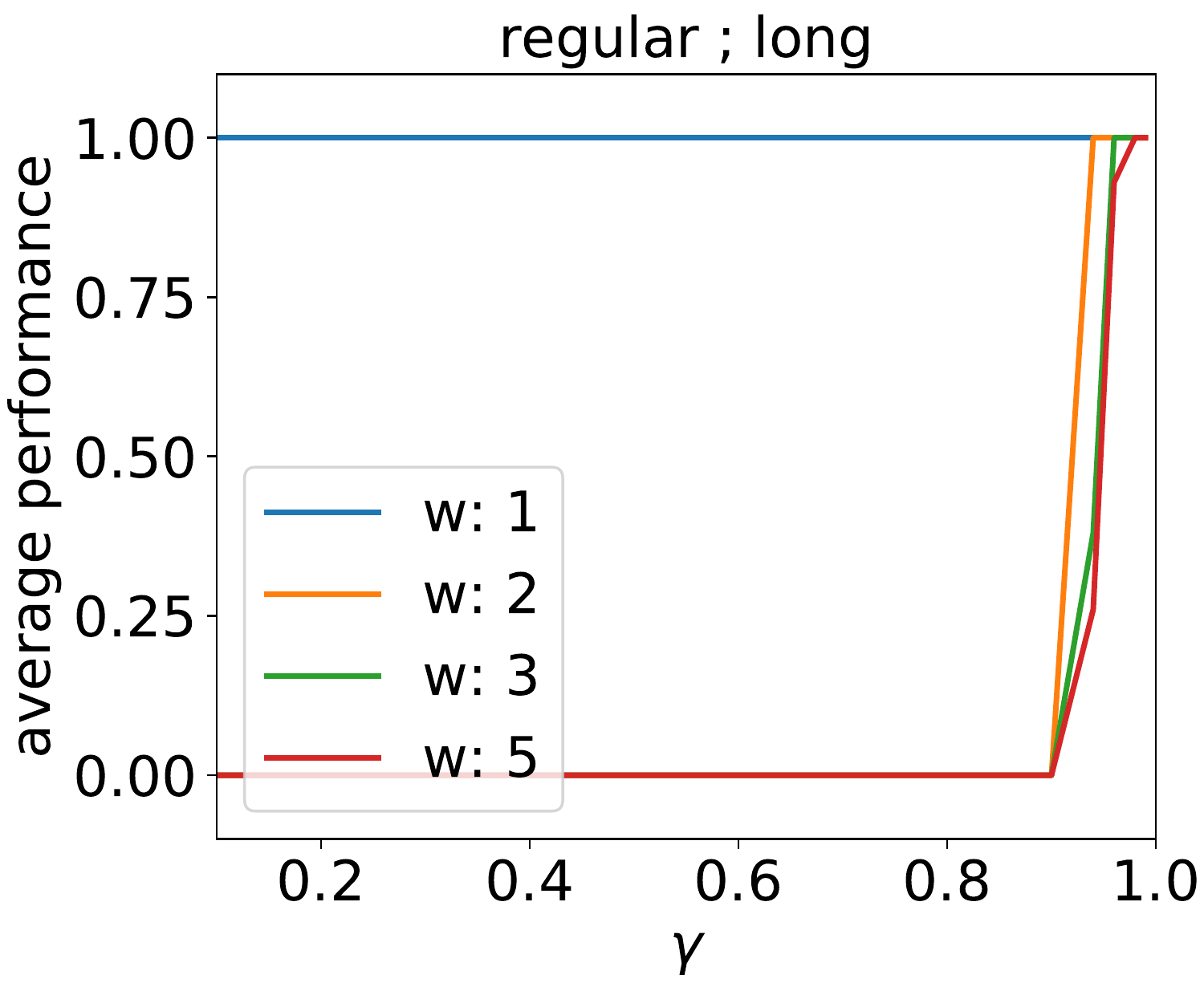}
\includegraphics[height=3.2cm]{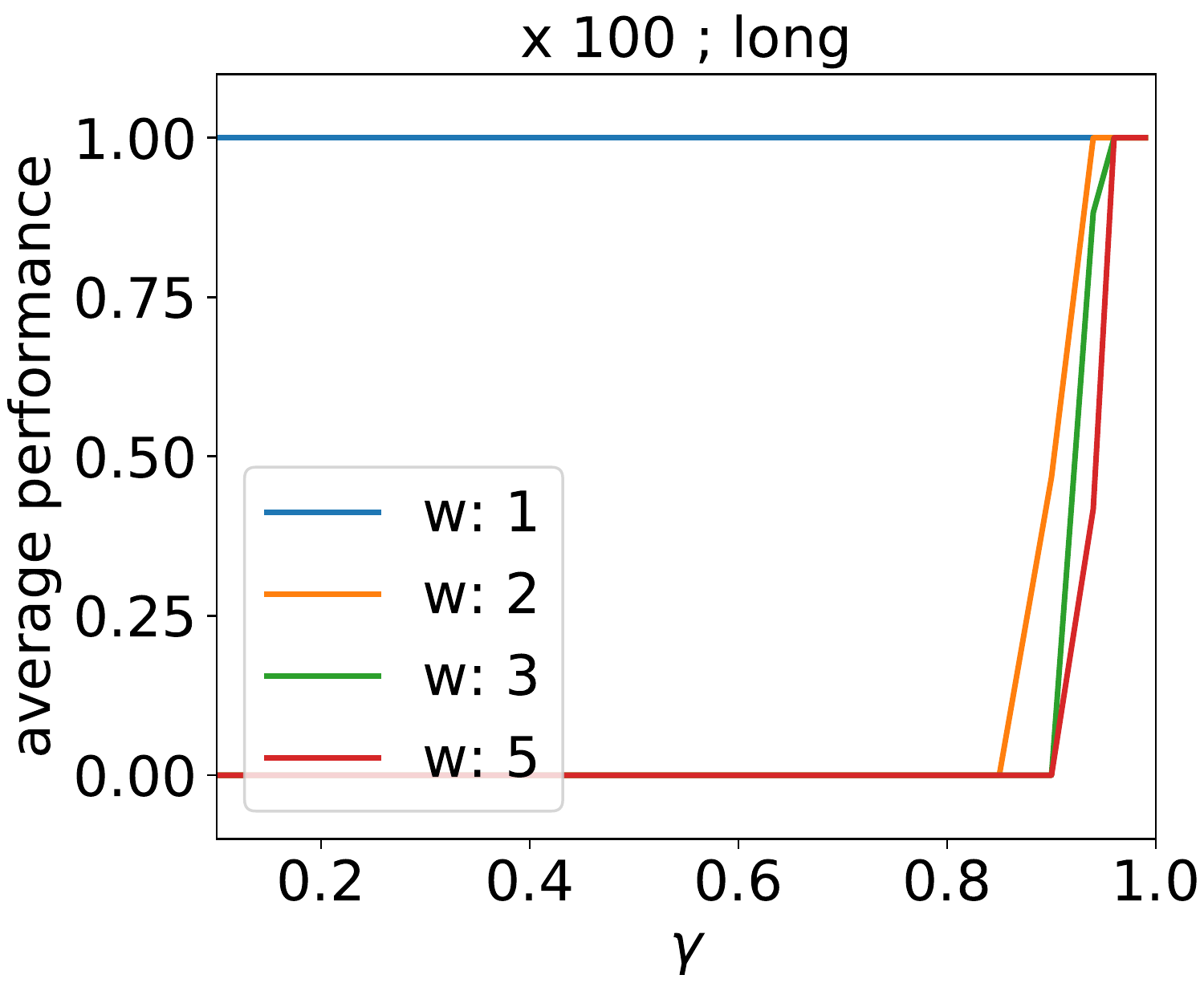}
\includegraphics[height=3.2cm]{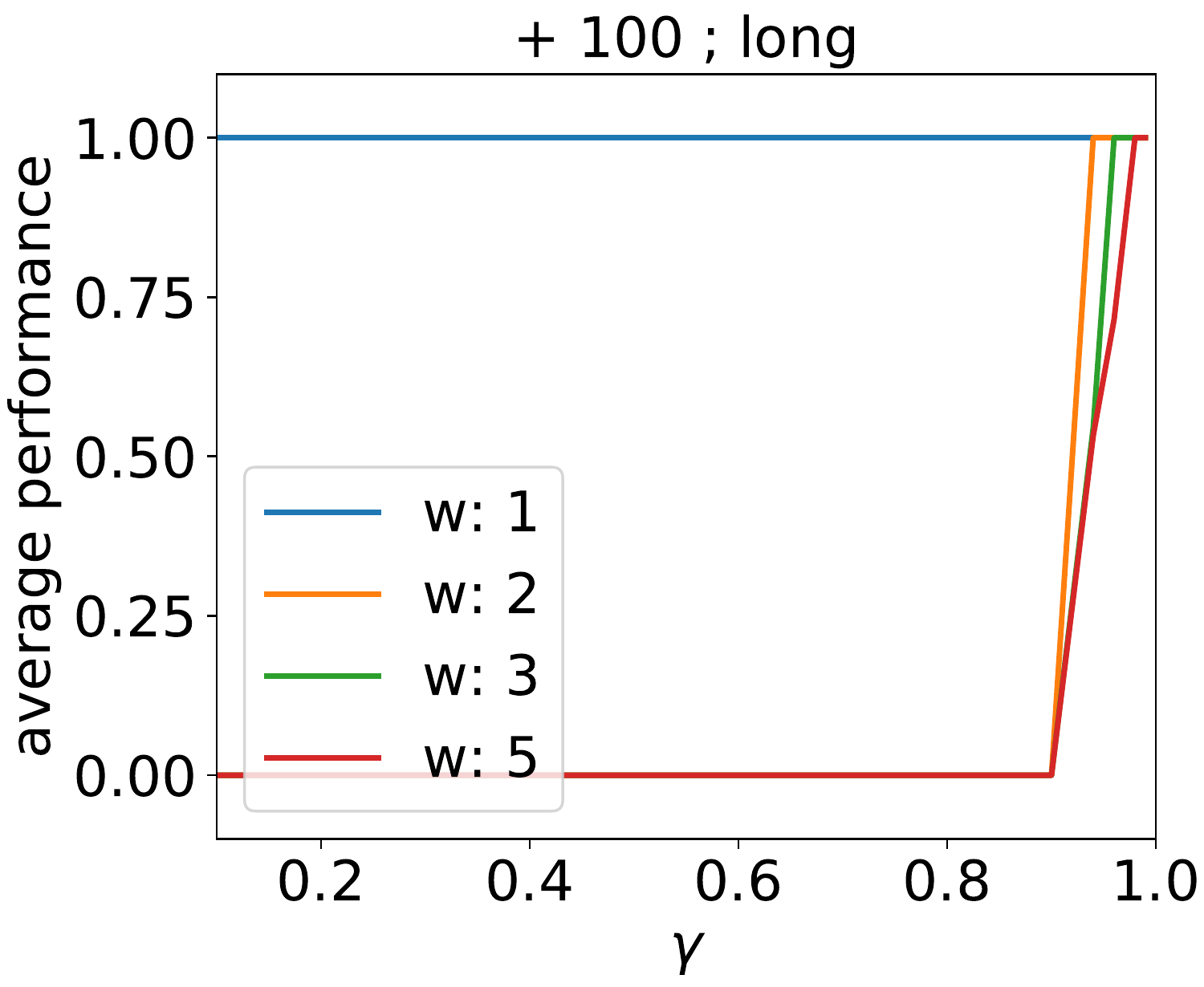}
\caption{Performance on 3 variations of the chain task. Left: performance on the regular version; middle: performance on the variant with values 100 times larger; right: performance on the variant with values pushed up by 100. All versions result in roughly the same performance curves.}
\label{fig:hypotheses testing}
\end{center}
\end{figure}

To test the first hypothesis, we performed the same experiment as in Section 3.2, but with rewards that are a factor $100$ larger. This in turn increases the action gaps by a factor $100$ as well. Hence, to validate the first hypothesis, this modification should improve (i.e., lower) the threshold value where the performance falls flat. To test the second hypothesis, we pushed all action-values up by $100$ through additional rewards, reducing the relative action-gap. Specifically, the extra reward upon transitioning to a non-terminal state $100\cdot(1-\gamma)$, while the extra reward upon transition to a terminal state is $100$. This effectively pushes all action-values up by exactly 100. To validate the second hypothesis, performance should degrade for this variation. We plotted the performance of these task variations, together with the performance on the regular task, in Figure~\ref{fig:hypotheses testing}. Both variations show roughly the same performance as the performance on the regular tasks, invalidating both hypotheses.

\section{Additional Details for Logarithmic DQN}

In this section, we describe additional details specific to our deep RL (Atari) experiments. 

\subsection{Implementation}

In order to support reproducibility and enable reliable and accessible baseline comparisons, we base our implementation upon the Google's Dopamine framework \citep{dopamine}. Dopamine provides reliable, open-source code for several important deep RL algorithms (including DQN) and enables standardized benchmarking, yielding `apples to apples' comparison under best known evaluation practices in RL. Therefore, we evaluate LogDQN without modifications of agent or environment parameters (w.r.t. those outlined by \citet{dopamine}), except for LogDQN's hyper-parameters (i.e., $\gamma, k, c, \beta_{log}, \beta_{reg}$, and $d$; for which the chosen values are stated in the paper).

We now highlight any settings in our LogDQN implementation that differs from our formulation of the logarithmic Q-learning update rules, as follows:

\begin{itemize}
    \item The most commonly-used loss function for DQN (and the default setting in Dopamine) is based on the Huber loss function \citep{huber1992robust}, which slightly differs from the squared-error loss specified as the general setting. While our results are for the standard Huber loss setting, in our primary experiments we did not observe any significant difference between the two.
    
    \item To optimize the loss function, we use the standard RMSProp optimizer\footnote{See \url{http://www.cs.toronto.edu/~tijmen/csc321/slides/lecture_slides_lec6.pdf}} (as the default setting in Dopamine's DQN). This choice differs slightly from our logarithmic Q-learning formulation which illustrates the case for the fundamental gradient descent method.
    
    \item To initialize the LogDQN network, we generally use the standard Xavier initialization \citep{xavier2010understanding} scheme (also a Dopamine's default setting), with the mere exception of initializing the output-layer weights of our $Q^-$ function to zero (instead of small, noisy values around zero). 
    
    \item We replaced the additive $\gamma^k$ in our original formulation of the mapping function, its inverse, and $d$ hyper-parameter with a minimum-clipping at $\gamma^k$ (i.e., enforcing the aforementioned value to be the minimum possible value in the corresponding computations). This gives a hard bound on the values that can be represented, instead of a soft bound, and increases the independence between the $k$ and $\gamma$ parameters, which is useful when optimizing hyper-parameters.
\end{itemize}

\subsection{Hyper-parameter tuning}
The hyper-parameters of LogDQN used for the experiments are the result from an earlier hyper-parameter optimization performed using an older version of LogDQN that did not have a strategy to deal with stochastic environments (as described in Section 4.2). Due to time-constraints, we were unable to perform a new hyper-parameter optimization for the full version of LogDQN.

\begin{figure}[tbh]
\begin{center}
\includegraphics[width=10cm]{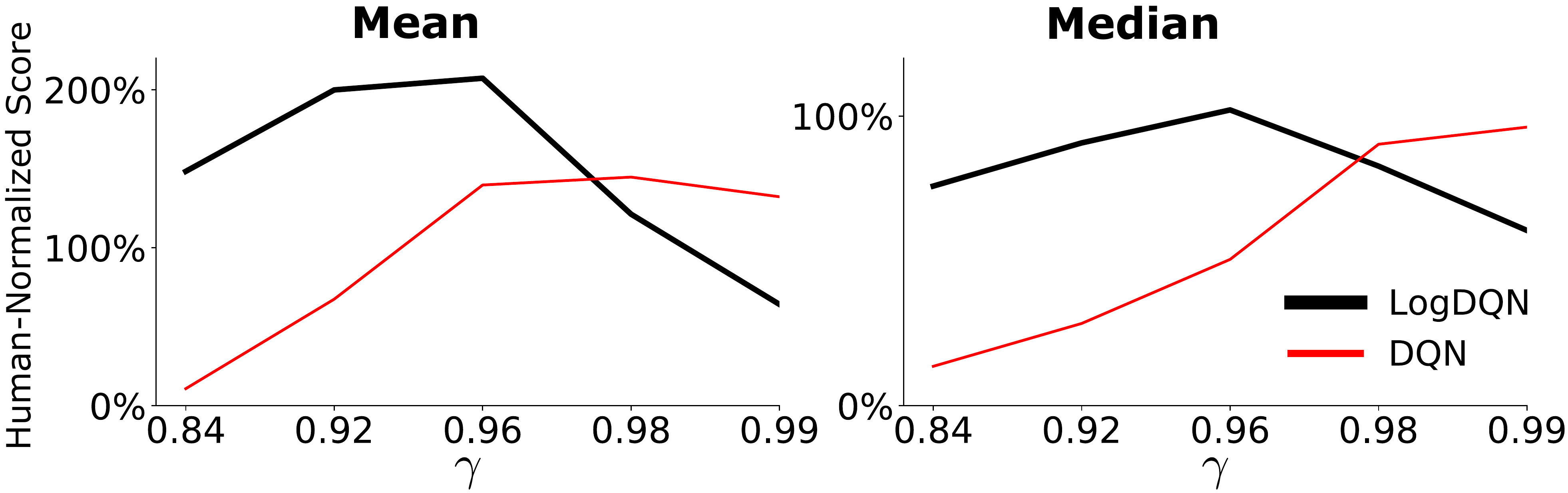}
\caption{Mean and median performance across 6 games for (an incomplete version of) LogDQN and DQN.}
\label{fig:mean_median_perf}
\end{center}
\end{figure}

This earlier hyper-parameter optimization was performed across these 6 games: \textsc{Alien}, \textsc{Zaxxon}, \textsc{Breakout}, \textsc{DoubleDunk}, \textsc{SpaceInvaders}, and \textsc{FishingDerby}. For the discount factor, we tried $\gamma \in \{0.84, 0.92, 0.96, 0.98, 0.99\}$ and for $c$ we tried $c \in \{0.1, 0.5, 1.0, 2.0, 5.0\}$. Furthermore, $k$ was fixed at 100. For DQN, we tried the same $\gamma$ values. Figure~\ref{fig:mean_median_perf} shows the mean and median human-normalized score across these 6 games.

In Figure~\ref{fig:mean_median_perf}, for LogDQN, the performance at the best $c$-value is plotted for each $\gamma$. The best values for LogDQN are $\gamma=0.96$ and $c=0.5$; for DQN, the best value is $\gamma=0.99$ (according to the more robust median metric).

\begin{figure}[tbh]
\begin{center}
\includegraphics[width=\columnwidth]{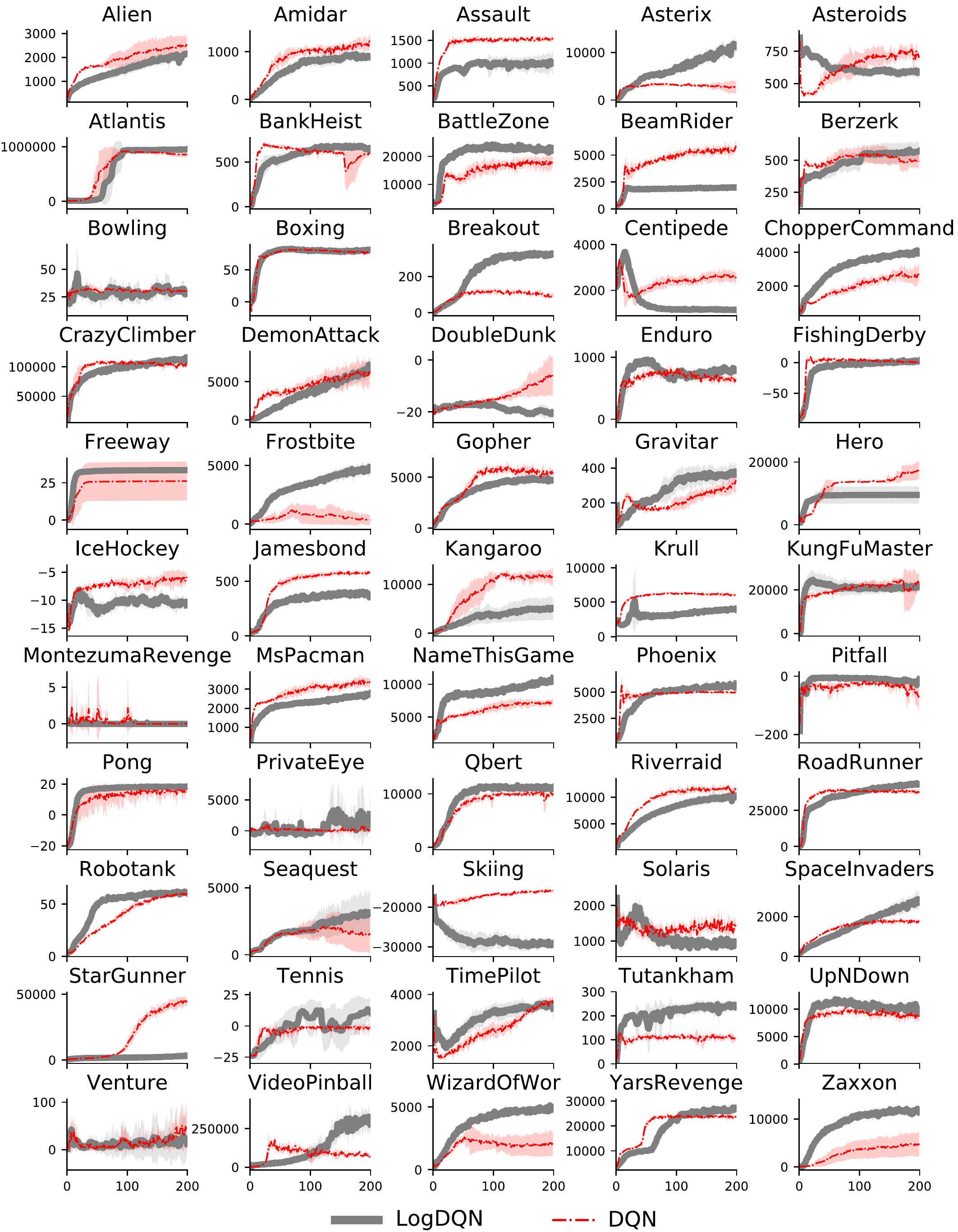}
\caption{Learning curves for all 55 games.}
\label{fig:all_learning_curves}
\end{center}
\end{figure}

\end{document}